\pdfoutput=1
\documentclass{article}

\usepackage{PRIMEarxiv}

\usepackage{siunitx}
\usepackage{comment}
\usepackage[normalem]{ulem}
\usepackage{xspace}
\usepackage{xcolor}

\DeclareSIUnit\px{px}
\DeclareSIUnit\fps{fps}
\DeclareSIUnit\dof{DoF}

\definecolor{OliveGreen}{RGB}{0,200,25}
\newcommand{\red}[1]{\textcolor{red}{#1}}

\newcommand{\darkgreen}[1]{\textcolor{OliveGreen}{#1}}
\newcommand{\blue}[1]{\textcolor{blue}{#1}}
\newcommand{\orange}[1]{\textcolor{orange}{#1}}

\newcommand{\ie}{i.\,e.\ }

\newcommand{\eg}{e.\,g.\ }
\newcommand{\Eg}{E.\,g.\ }

\newcommand{\armarIII}{\mbox{ARMAR-III}\xspace}

\newcommand{\armarx}{\mbox{ArmarX}\xspace}

\newcommand{\ackReallaborTERRINet}{The research leading to these results has received funding from the Baden-Württemberg Ministry of Science, Research and the Arts (MWK) as part of the state's "digital@bw" digitization strategy in the context of the Real-World Lab "Robotics AI" and  from the European
Union’s Horizon 2020 research and innovation programme under the grant1220
agreement No 730994 (TERRINet).}

\newcommand{\added}[1]{\darkgreen{#1}} %\newcommand{\added}[1]{\blue{#1}}
\newcommand{\replaced}[2]{\red{\ifmmode\text{\sout{\ensuremath{#1}}}\else\sout{#1}\fi}\darkgreen{#2}}
\newcommand{\removed}[1]{\red{\ifmmode\text{\sout{\ensuremath{#1}}}\else\sout{#1}\fi}}
\newcommand{\remark}[1]{\blue{#1}}

\newcommand{\todo}[1]{\{\orange{---TODO--- #1}\}}

\newif\iffinal
\iffinal
	\renewcommand{\added}[1]{#1}
	\renewcommand{\replaced}[2]{#2}
	\renewcommand{\replaced}[2]{\added{#2}}
	\renewcommand{\removed}[1]{}
	\renewcommand{\remark}[1]{}
	\renewcommand{\todo}[1]{}
\fi
\newcommand{\removedfootnote}[1]{\footnote{\removed{#1}}}
\newcommand{\removedsubsection}[1]{\subsection{\texorpdfstring{\removed{#1}}{#1}}}

\title{A Memory System of a Robot Cognitive Architecture and its Implementation in \armarx}
\author{
  \textbf{Fabian~Peller-Konrad, Rainer~Kartmann, Christian R.\,G. Dreher,}\\ \textbf{Andre~Meixner, Fabian~Reister, Markus~Grotz and Tamim~Asfour}\\
  High Performance Humanoid Technologies (H$^2$T) \\
  Institute of Anthropomatics and Robotics (IAR) \\
  Karlsruhe Institute of Technology (KIT) \\
  Karlsruhe, Germany\\
  \texttt{\{fabian.peller-konrad, rainer.kartmann, c.dreher,} \\\texttt{andre.meixner, fabian.reister, markus.grotz, asfour\}@kit.edu}
}
\newcommand{\citet}[1]{\cite{#1}}
\newcommand{\citep}[1]{\cite{#1}}
\usepackage{url}
\usepackage{lineno}
\usepackage{microtype}
\usepackage[onehalfspacing]{setspace}

\usepackage{pdfpages}
\usepackage{booktabs}
\usepackage{array}
\usepackage{tabularx}
\usepackage{makecell}

\usepackage{algorithm}
\usepackage[noend]{algpseudocode}
\usepackage{placeins}

\usepackage{hyperref}

\usepackage{listings}
\usepackage[inline]{enumitem}
\usepackage{color}
\definecolor{gray}{rgb}{0.4,0.4,0.4}
\definecolor{darkblue}{rgb}{0.0,0.0,0.6}
\definecolor{cyan}{rgb}{0.0,0.6,0.6}

\usepackage{cleveref}
\usepackage{rotating}
\usepackage{relsize}
\usepackage{siunitx}

\def\cpp{\texorpdfstring{\protect C\nobreak\hspace{-.05em}\raisebox{.4ex}{\relsize{-3}\textbf{++}\xspace}}{C++}}
\newcommand{\python}{Python\xspace}

\crefname{algorithm}{Algorithm}{Algorithms}
\crefname{figure}{Figure}{Figures}
\crefname{table}{Table}{Tables}
\crefname{section}{Section}{Sections}
\crefname{subsection}{Section}{Sections}

\usepackage{amssymb}% http://ctan.org/pkg/amssymb
\usepackage{pifont}% http://ctan.org/pkg/pifont
\newcommand{\cmark}{\ding{51}}%
\newcommand{\xmark}{\ding{55}}%

\begin{document}
\onecolumn
% if frontiers
%\firstpage{1}

\maketitle

\begin{abstract}

Cognitive agents such as humans and robots perceive their environment through an abundance of sensors producing streams of data that need to be processed to generate intelligent behavior. A key question of cognition-enabled and AI-driven robotics is how to organize and manage knowledge efficiently in a  cognitive robot control architecture. We argue, that memory is a central active component of such architectures that mediates between semantic and sensorimotor representations, orchestrates the flow of data streams and events between different processes and provides the components of a cognitive architecture with data-driven services for the abstraction of semantics
from sensorimotor data, the parametrization of symbolic plans for execution and prediction of action effects.

Based on related work, and the experience gained in developing our ARMAR humanoid robot systems, we identified  conceptual and  technical requirements of a memory system as central component of cognitive robot control architecture that facilitate the realization of high-level cognitive abilities such as explaining, reasoning, prospection, simulation and augmentation. Conceptually, a memory should be active, support multi-modal data representations, associate knowledge, be introspective, and have an inherently episodic structure. Technically, the  memory should support a distributed design, be access-efficient and capable of long-term data storage.
We introduce the memory system for our cognitive robot control architecture and its implementation in the robot software framework \armarx.  
We evaluate the efficiency of the memory system with respect to transfer speeds, compression, reproduction and prediction capabilities.

\tiny

 % FRONTIERS
 %\keyFont{ \section{Keywords:} } %All article types: you may provide up to 8 keywords; at least 5 are mandatory.
\end{abstract}

\section{Introduction}

%TODO Klar machen, dass wir ,it "data", "information" und "knowledge" üblicherwise das gleiche meinen?

% Desired characteristics for a cognitive architectures by Vernon:
% perception, action, learning, adaptation, anticipation, attention, action-selection, motivation, autonomy, internal-simulation, memory, reasoning, meta-reasoning/meta-cognition

% We believe that memory should support cognitive processes. These are:
% action, perception -> multi-modal
% learning -> episodic, introspective, associative
% anticipation -> episodic, introspective, associative
% adaptation -> active, meta-cognition, 
% reasoning -> introspective, associative
% attention -> active, introspective
% motivation -> active, introspective
% internal-simulation -> introspective, augmentation
% 

\label{sec:introduction}

%Categorization of perception-action dependencies as non symbolic object-action-effect representations

%Generative modelling and symbolic representation to bootstrap generalized and transferable symbolic knowledge of perception-action associations and dependencies (i.e. Know-how) from specific perception-action contingencies of objects, associated actions and consequent effects

%Internal simulation to predict the outcome of actions or plan actions to achieve desired outcomes

%Enactive grounding to validate internal simulation results by observing the actual result and either reinforcing or weakening the generative model that produced them

%--------------------------------------------------------------

\begin{figure}[bht]
    \centering
    \includegraphics[width=\linewidth, bb=0 0 1000 500,clip]{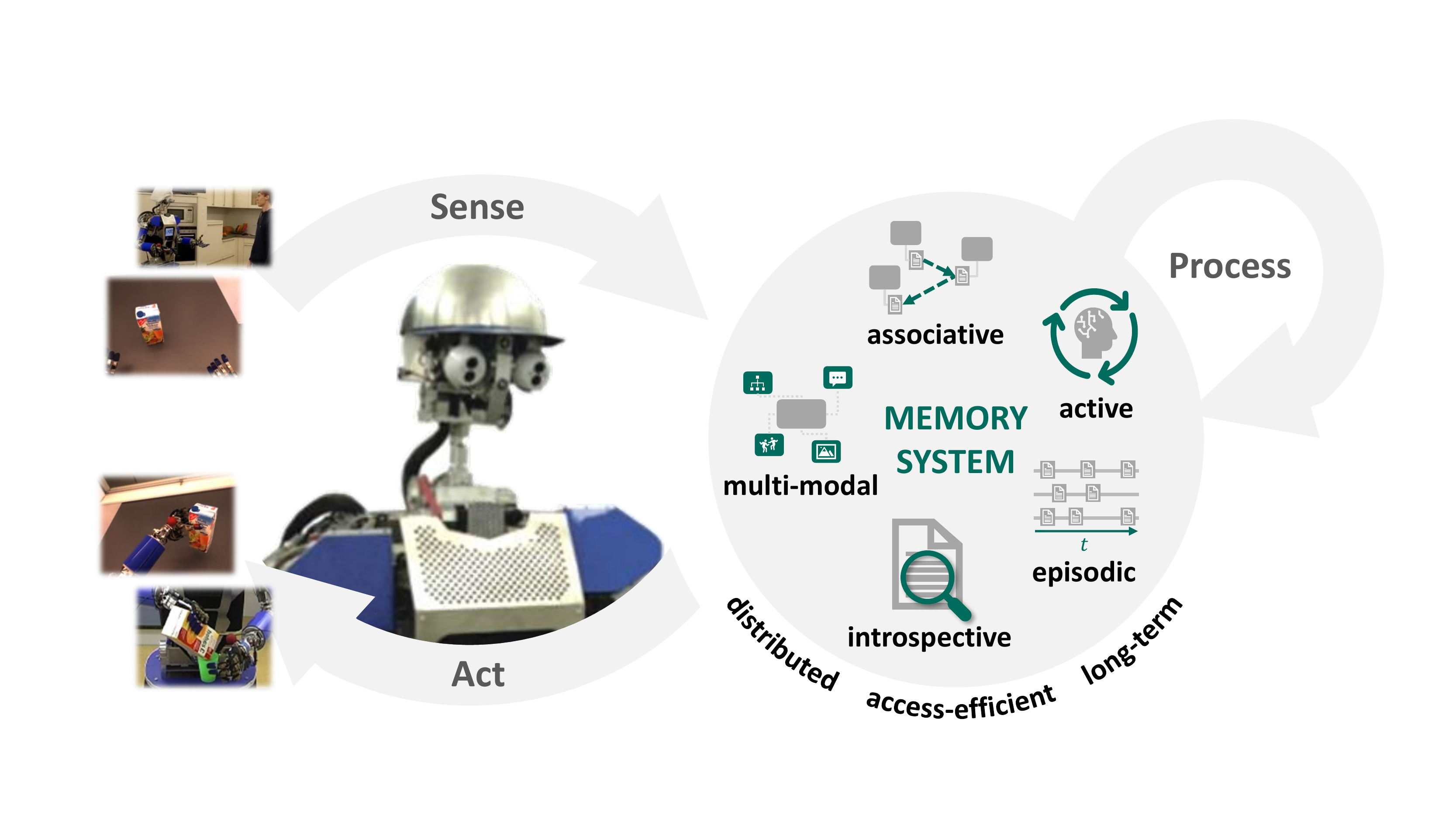}
    \caption{The humanoid robot \armarIII~\cite{ARMAR3} perceiving multi-modal data of different sensors such as haptic, visual and audio information. This data has to be stored and processed in an adequate memory system. To optimally support various cognitive processes and action, a memory system must fulfill several characteristics, such as an active design, multi-modality, an inherently episodic structure, associativity and introspectability.}
    \label{fig:motivation}
\end{figure}

% Motivate with the memory system in a human
The human memory is one of the most astonishing and intricate systems in nature as it receives a huge amount of sensory data and processes it in a highly distributed way~\cite{Wood_2012}.
%  Explain why the concept of a human is interesting for implementing robots - should be shorter!
Humanoid robots face similar challenges, thus, developing memory-based cognitive control architectures is a key challenge to endow such robots with intelligent behavior and seamless human-like interaction abilities with the environment.
A humanoid robot has multiple sensors producing sensor data and actuators executing motor commands which in turn allow performing actions as depicted in \cref{fig:motivation}.
Several processes are needed for filtering, interpreting, and using the data to perform tasks such as object detection, task planning and execution, navigation, motion generation and control, learning and interaction.

% Design of a cognitive system in a robotics architecture
How can we create humanoid robot systems with rich cognitive and sensorimotor capabilities that  are comparable to the human's, especially regarding learning and development?
The authors in \cite{Vernon_2014, Vernon_2014_CV} describe cognition as ``the process by which an autonomous system perceives its environment, learns from experience, anticipates the outcome of events, acts to pursue goals, and adapts to changing circumstances''. 
This definition identifies three core components for robotics. A cognitive system needs (i) \emph{perception} components in order to perceive the environment, (ii) \emph{processing} components in order to learn from experience, anticipate the outcome of actions and adapt to changing circumstances, and (iii) \emph{action execution} components in order to purposely act to achieve goals.
In addition, a robot's cognitive architecture requires a place  to hold information acquired from perception and action execution so that the  processing components can access and process the data and eventually store the results back in this place -- the memory.  
We argue that a \emph{memory system} (or \emph{memory} in short) is a key element in any cognitive architecture. Such a memory should not only connect system components and store multi-modal information on different levels of abstraction but provide mechanisms and services for abstraction of semantics from sensorimotor data, parametrization of symbolic plans for execution in a given context and prediction of action effects.  

% Introduce the memory + signal to symbol gap
To this end, we formulate our first hypothesis:
\textit{A memory system in a cognitive robot control architecture mediates between i) high-level abilities, usually represented in a symbolic manner, such as language understanding, scene understanding, planning, plan execution monitoring and reasoning, and ii)  low-level abilities, such as sensor data processing, sensorimotor control.} 
%We locate abilities such as language and scene understanding, planning, reasoning or plan execution monitoring on the highest layer. Sensorimotor control and sensor data processing are part of the lowest layer.
This means that the memory system must be able to process a huge amount of data  -- no matter if the data is symbolic (\eg plans, words, relations, etc.) or sub-symbolic (\eg images, joint configurations, forces, etc.). It must build a bridge between sub-symbolic (sensorimotor) representation  and symbolic (semantic) representation, tackling the \emph{signal-to-symbol-gap} by \eg learning representations of perception-action dependencies in form of Object-Action Complexes (OACs) as proposed by~\citep{Kruger2011OACs}.  
%as well as makes the occurring data available for access by the agent's cognition.  

This observation leads to our second hypothesis: 
\textit{Multi-modal representations are key.}
The ability to store multi-modal information, the efficiency of storage and retrieval or the ability to learn from such data require a meaningful and efficient multi-modal representation of knowledge. This representation must be specific enough to differentiate between symbolic and sub-symbolic information. On the other hand, the representation must also support generalization on sensorimotor and symbolic level. Further, it must support \emph{associations of knowledge}~\citep{Wood_2012} because the system needs an understanding of how perception and action are coupled and which sensations usually occur together. Such multi-modality applies explicitly to  information resulting from different cognitive processes within the architecture. 
 
\begin{figure}[ht]
    \centering
    \includegraphics[width=\linewidth, bb=0 300 600 800,clip]{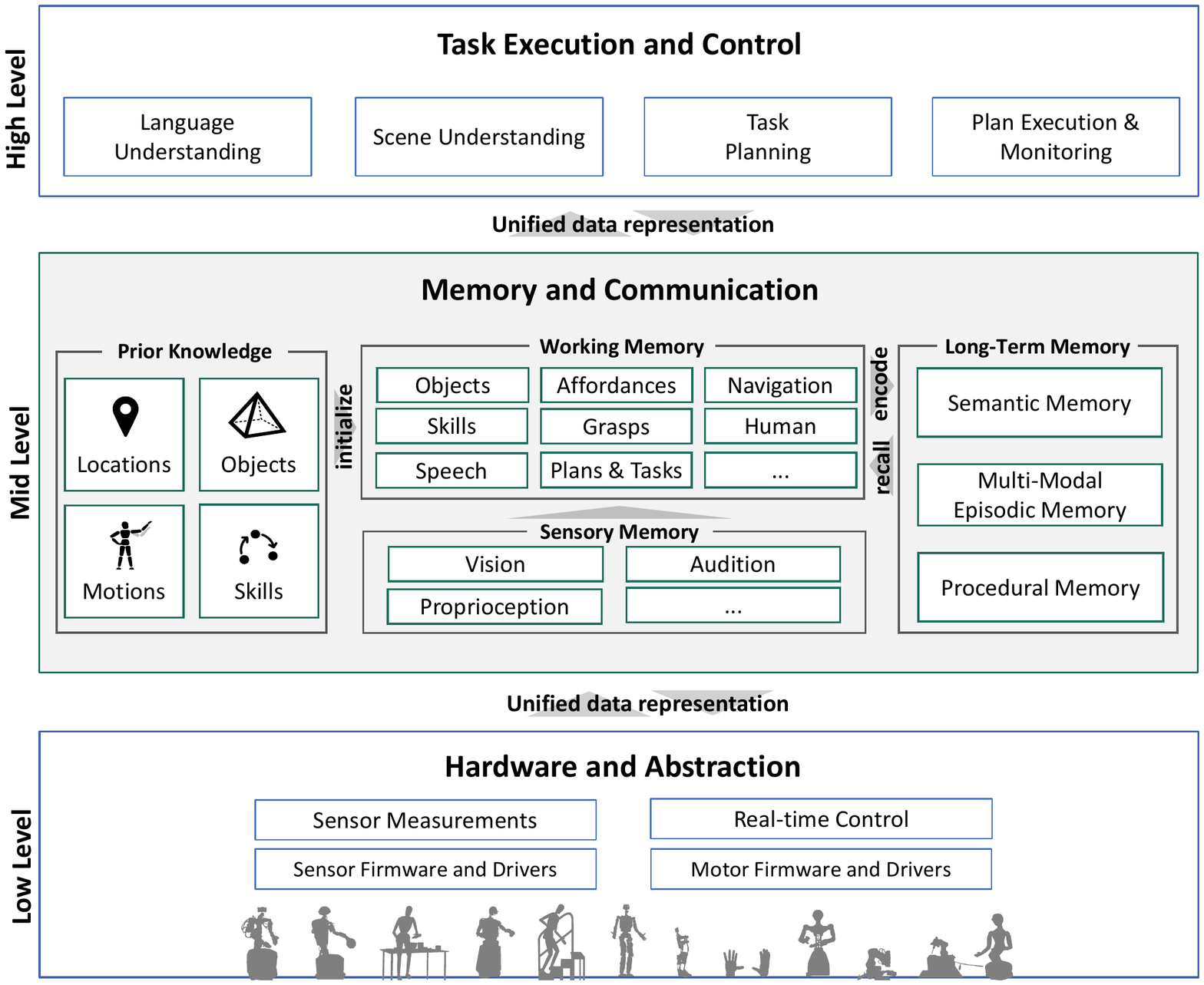}
    \caption{Overview of the memory system as mediator between a symbolic high level and sub-symbolic low level. The memory system is implemented in the robot software framework \armarx \protect\footnotemark}
    \label{fig:armarx:architecture}
\end{figure}
\footnotetext{See \url{https://armarx.humanoids.kit.edu} and~\cite{Vahrenkamp_2015}.}

The main contributions of this paper are two-fold: (i) On a conceptual level, we identified requirements (or characteristics) of memory systems of a cognitive control architecture of a humanoid robot to support our two hypotheses. These requirements comprise the five concepts \emph{active, episodic, multi-modal, associative and introspective}, as well as the three technical requirements \emph{distributed, access-efficient and long-term}.  We argue that a memory is an active element within the architecture, \ie it is not simply a passive storage device but it is able to process multi-modal, both symbolic and sub-symbolic data. The ability to inspect information and to adapt behaviors based on such information greatly supports the active role on the memory. Further, we argue that it is beneficial to manage data episodically, no matter if it is declarative or non-declarative knowledge. Finally, we explain that associations of how knowledge is shared within the system and how such knowledge is extracted from data yield great benefits for data processing, learning and development. The technical requirements are all related to improving the system's overall efficiency. Further, we promote the need for memory-driven cognitive control architectures for complex robot systems such as humanoid robots with these core characteristics to create  systems with advanced cognitive abilities. (ii)  The above considerations served as guidelines for the implementation of the memory system with the described core characteristics in our robot software framework \armarx.

%efficient, scalable, flexible, yet understandable and to derive important cognitive abilities, such as explainability, reasonability, simulation or predictability.  \armarx\footnote{\url{https://armarx.humanoids.kit.edu}}~\cite{Vahrenkamp_2015}.
We take inspiration from cognitive science \cite{Squire_2004} for the realization of the memory system of our cognitive robot control architecture but with special focus  on optimizing data flow and making the development and integration of new components easier. 
%, is the central part of our cognitive robot control architecture as shown in \cref{fig:armarx:architecture}. It acts as a mediator between low and high-level components. 
The implementation further includes a novel way for representation information in the memory, which allows the inspection of information at run-time.
Overall, we provide an overview of the technical implementation of the memory system that meets the requirements of complex robotic systems, such as humanoid robots, including a broad evaluation of its efficiency.
We would like to emphasize that with this work we do not describe a complete cognitive architecture, but focus on the aspect of a memory system in such architectures.  
From a user perspective, we show how our new memory system optimizes data flow, enables mediation between high- and low-level in operation and leverages the integrated key characteristics in multiple use cases.

The remainder of this paper is structured as follows:
\cref{sec:related_work} gives an overview of the memory classes known in cognitive psychology, which are often used for the realization of memory-based cognitive architectures for technical systems. 
In addition, we compare related cognitive architectures with a special focus on their design and implementation of the memory including a comparison of the conceptual characteristics.
The conceptual and technical concepts of our memory system are then motivated and described in detail in \cref{sec:requirements}. Subsequently, we provide a technical overview of our memory system and its implementation in \armarx, including the working memory, the data representation format and the long-term memory in section \cref{sec:implementation}
A broad quantitative evaluation of our memory system follows in \cref{sec:evaluation}.
\cref{sec:casestudies} discusses use cases of the memory system, demonstrating the usage of its new capabilities based on the implemented characteristics. 
In \cref{sec:conclusion} the contributions of this work are summarized and planned extensions of the presented architecture are discussed.

%This decision accounts for the fact that an agent's semantic and procedural knowledge is subject to change when new experiences are made from which general knowledge about the world is derived.
%For instance, 
%consider an object learning algorithm deriving, due to limited observations, that an \emph{apple} is a \emph{table}.
%At a later point in time, it might get new data and correct its previous decision.
%In such cases, it is essential to not just keep the most recent knowledge, 
%but to also know when and why the knowledge has been updated 
%in order to improve the performance of the learning algorithm for the future.

\section{Related Work}
\label{sec:related_work}

In this section, we briefly describe the different memory structures as motivated from cognitive psychology.
%, without going much into detail about the neuro-scientific aspects of a memory.
Second, we discuss related cognitive architectures
with respect to the identified core characteristics of a memory system.

\subsection{Memory}

\begin{figure}
    \centering
    \includegraphics[width=\linewidth,bb=0 100 800 450,clip]{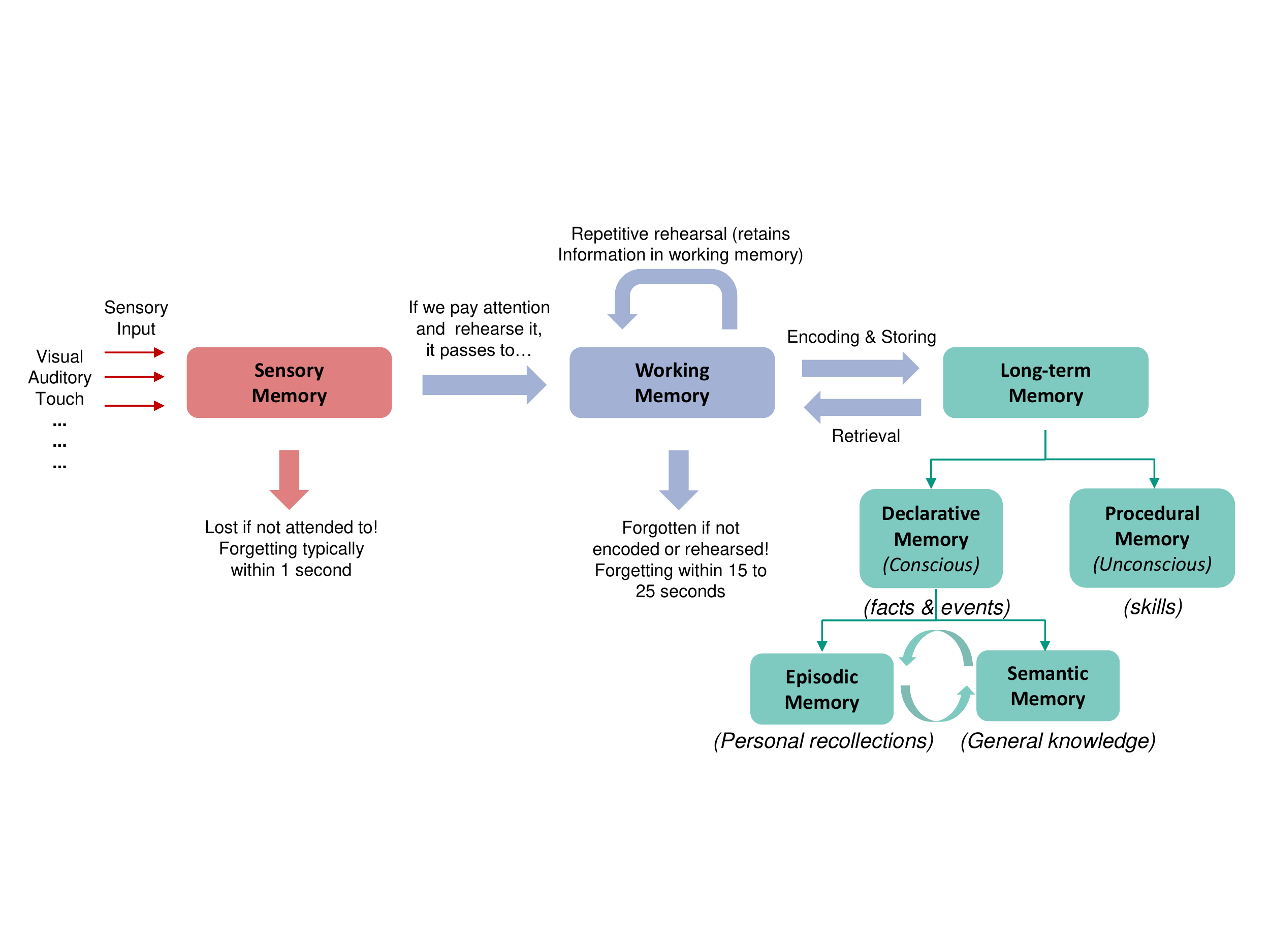}
    \caption{A taxonomy of memory classes. Multi-modal sensory information is passed from a sensory memory over a working memory to a long-term memory. Long-term memory itself is be subdivided into declarative memory and non-declarative (here procedural) memory.}
    \label{fig:memory_taxonomy}
\end{figure}

By studying amnesic patients and animals in the second half of the $20$th century, researchers found out that different areas of the brain are responsible for different memory tasks, motivating the assumption that memory consists of several subsystems~\cite{Squire_2004}. Findings from the medical history of famous patients such as \emph{Henry Gustav Molaison} or \emph{Clive Wearing} greatly influenced the development of cognitive psychology and theories that attempt to explain the connection between brain function and memory~\cite{HM_CW}. 

Based on such theories of memory,~\cite{Atkinson_1968} proposed the so-called \emph{Multi-Store Model}. This theoretical model of human memory introduced three types of memory:
\begin{enumerate*}[label=(\roman*)]
\item A sensory memory that processes perceptual information, 
\item  a long-term memory that holds information for a long duration, and \item a short-term memory that holds information through repetitive rehearsal and which receives information from the sensory memory through attention and from long-term memory through retrieval. 
\end{enumerate*}
Newer works distinguish between short-term memory and working memory or question a clear distinction between different memory types for biological systems altogether~\cite{aben2012}. However, such taxonomies provide a basis to structure and classify artificial memory systems into a taxonomy. 
\cref{fig:memory_taxonomy} shows a taxonomy of memory classes. Muli-modal sensory information is passed from a sensory memory over a working memory to a long-term memory. Long-term memory itself is be subdivided into declarative memory and non-declarative (here procedural) memory.

\subsubsection{Sensory Memory}
The sensory memory (SM) holds perceptual information for a short time duration and thus it acts as a cache memory or repository for incoming sensory information. 
In biological systems, sensory receptors take \eg visual, auditory or touch information and forward it directly to the nervous system for further processing. It is common to distinguish between different sensory memory types for different modalities such as iconic memory for visual information, echoic memory for auditory information, etc.~\cite{Wan2020}.
%Iconic memory refers to the visual information stored in this this short-term cache and it is thought to hold information for less than a second. Aural information is stored in a so called echoic memory. The echoic memory is assumed to have a longer decay rate than the iconic memory~\cite{Vernon_2021}.
%Information related to touch interactions but also proprioceptive information (e.g. joint angles and relative position of the body) are stored in a so called haptic memory in which information decays after around two seconds.
The sensory memory is assumed to be a highly volatile storage containing raw, unanalyzed sensory data that is derived from the senses. The data is only stored long enough to be passed to the working memory (WM). To limit the amount of data transferred to the WM, the data is only transferred when it is attended to, and is lost otherwise.

\subsubsection{Working Memory}
Similar to the SM, the working memory (WM) holds information for a limited time duration. However, this duration is longer and is in the order of seconds~\cite{Cowan1997, Cowan_2001}. In addition, the WM can consciously be controlled by attention mechanisms and is therefore important for reasoning, learning, problem-solving, and other mental processes. The ability to have certain details ready, even if they are not yet stored in long-term memory, supports a variety of everyday mental processes.  Examples include remembering the first part of a sentence or keeping a number in mind while solving a mathematical problem. Studies show that people are able to keep track of several items at the same time~\cite{Barak_2014}. This motivates the assumption that the WM has a relatively small capacity of $7 \pm 2$ \emph{chunks}~\cite{Miller_1956}, when not using exploits like repeating information out loud, regardless of whether the elements are digits, letters, words, or other units. Newer research distinguish between the modalities,\eg seven chunks for digits, six for letters, and five for words~\cite{Service_1998}. 
%\citet{Cowan_2001} proposed that the WM only has a capacity of about four chunks in young adults. The complexity of information stored in each chunk differs from person to person. While most adults can only repeat about seven digits correctly, some people have shown impressive improvements of being able to repeat up to 80 digits. This improvement can be achieved through massive training, \eg learning to improve the way data is grouped together and how these groups are encoded into individual units. 
According to~\cite{Gobet_Simon_2000}, the total amount of chunks can not be increased -- we can only increase the complexity of the referred information. The WM is not only exclusive to humans as animals have also shown similar abilities such as storing and maintaining several items simultaneously in memory, remembering their order and manipulating them~\cite{Barak_2014}. 

% Drin lassen für rev1?
The working memory is sometimes used interchangeably with short-term memory (STM)~\cite{aben2012}, but some consider the two forms of memory to be distinct and argue that working memory allows for the manipulation of stored information, whereas short-term memory refers only to the short-term storage of information~\cite{Baddeley1974}. For simplicity, we do not explicitly distinguish between WM and STM in this work.

\subsubsection{Long-Term Memory}
The long-term memory (LTM) is intended for the storage of information over a long period of time. 
%Some people even assume that the human LTM is able to store information infinitely and one only forgets how to recall that information \cite{}. 
Through the process of repetitive rehearsal and association, memories of WM consolidate to LTM or an existing memory in the LTM gets reinforced~\cite{Atkinson_1968}. The WM can then retrieve the data from the LTM when it is needed for processing. %While in short-term memory, the memory is maintained by transitory electrical activity while long-term memory uses longer-lasting chemical changes in the neural system. 
During the process of consolidation, data is encoded into a special representation to identify groups and for generalization. However, memories stored in LTM are not saved in a static state. % \cite{https://www.verywellmind.com/what-is-long-term-memory-2795347}. 
Studies showed that memories in LTM are transformed every single time they are accessed~\cite{taylor1983}.
According to~\cite{Squire_2004}, the LTM can be divided into two types.
\begin{enumerate*}[label=(\roman*)]
\item \emph{Declarative memory}, also known as explicit memory, that contains information such as facts and events and is managed through conscious control, and 
\item \emph{Non-declarative memory} contains implicit knowledge, such as the ability to perform various actions or behavioral control parameters.
%There is evidence that both, the LTM and the WM hold episodic and semantic information although most taxonomies of the memory only 
\end{enumerate*}
\ \\ 

\emph{Declarative Memory.}

%Declarative Memory is the kind of memory that is meant when the term ``memory'' is used in everyday language.
The declarative memory can be further subdivided into two types. The \emph{semantic memory} consists of general knowledge about the internal state and the environment such as facts, ideas and concepts. In the context of cognitive architectures for technical systems, knowledge stored in this part of the memory is usually assumed to be symbolic~\cite{Kotseruba_Tsotsos_2017}. In comparison, the \emph{episodic memory} contains episodes or autobiographical experiences occurring in an explicit spatial and temporal context, \ie what, where, and when something happened. 
There is evidence that the knowledge of how information has changed in the past in the episodic memory as well as the knowledge about facts in the semantic memory form the basis for prediction and explanation. There is a strong coupling between them, as we derive new concepts from the experiences we have stored in the episodic memory~\cite{graham2000}.
\ \\ \

\emph{Non-Declarative Memory.}

The non-declarative memory covers all information that is not consciously accessible. Again, this memory can be further subdivided.
\emph{Priming} describes the ability to strongly accelerate the retrieval of information from long-term memory by a related stimulus. It requires that knowledge can be associated heterogeneously, \ie no matter how the knowledge is represented. Priming can be further subdivided into positive, negative, semantic, perceptual and conceptual priming.
In \emph{classical conditioning}, different stimuli are linked together. A well-known example of classical conditioning is \emph{Pavlov's dog}~\cite{coon2018}, which showed increased saliva production just by ringing the dinner bell. The neutral stimulus ``ringing of the bell'' was thus linked with the positive stimulus ``there is food'', which triggers a physical reaction. \emph{Non-associative learning} is the simplest form of learning as it does not require stimuli association~\cite{Ioannou2018}. Habituation and sensitization are the two forms of non-associative learning. Habituation describes the process of inhibiting a response after repeated exposures to a stimulus. The degree to which a response is inhibited depends on the repetition rate of the stimulus, its intensity, the duration of the stimulus, and how often the agent is exposed to the stimulus. On the other hand, a sensitized stimulus has increased intensity and sensitization does not require repetitive stimuli. Even a single stimulus may cause a reinforced response. For example, relapse of addiction can be seen as sensitization. Even a few stimuli, \eg from drugs or gambling, can trigger a strong physical desire. The \emph{procedural memory} is required for skilled behaviors and habits. There is less known about how humans store skills and abilities except that skills are learned and refined through practical training and that for learning a skill a large variety of areas of the brain are involved.

From the non-declarative memory, artificial systems often only explicitly implement procedural memory~\cite{Kotseruba_Tsotsos_2017}. However few approaches focus on how artificial systems can be conditioned and how this mechanism can be used in social robots~\cite{Novianto2014}.

%\subsubsection{Memory properties}

% Given the taxonomy in \ref{fig:memory_taxonomy} 

%\subsubsection*{Modality}

%\subsubsection*{Representation}

%\subsubsection*{Associativity}

%\subsubsection{Finite and Infinite Memories}

\subsection{Artificial Cognitive Architectures}
\label{sec:cognitive_architectures}

The development of artificial cognitive architectures is a longstanding and still unsolved problem, \eg the development of the cognitive architecture \textit{ACT-R}~\citep{Anderson2004} started in the early 1980s and is still ongoing. % as there is no universal and all-encompassing definition of cognition. 
In general, artificial cognitive architectures try to explain and represent the underlying processes of human cognition, such as perception, attention, action selection, memory, learning, reasoning, meta-cognition or prospection~\cite{Vernon_2021}. There is no implementation that is able to solve all these tasks with similar performance to humans~\cite{Kotseruba_Tsotsos_2017}. Thus, the proposed architectures focus on different aspects of cognition.

% \pimpladapted{
% Regarding the knowledge representation especially for embodied cognitive architectures~\cite{Paulius_2019} identified six characteristics for a  general knowledge representation that encompasses all of the necessary information needed by an agent to solve a problem and understand its environment: (i) it must be able to represent motions for actions, (ii) it must be able to represent perceptual data, (iii) it must grounded in logic, (iv) it must be able to model probabilities and beliefs, (v) it must support learning from experience, (vi) it must support a descriptive problem statement. 
% }
\cite{Kotseruba_Tsotsos_2017} estimate the number of artificial cognitive architectures to be around three hundred, of which about one hundred are being actively developed. 
Artificial cognitive architectures can roughly be divided into three classes \cite{Kotseruba_Tsotsos_2017, Vernon_2007, Vernon_2014, Sun_2004, duch2008}:
\begin{enumerate*}[label=(\roman*)]
    \item \emph{Cognitivistic architectures}, where knowledge is usually represented as symbols, 
    \item  \emph{emergent architectures} that focus on sub-symbolic processing and self-supervised learning of how knowledge is associated, and 
    \item \emph{hybrid architectures} that combine both types of processing. 
\end{enumerate*} 
Not every cognitive architecture is fully integrated in the sense that it covers all the aforementioned processes of human cognition.
%\pnew{
%In the following we will describe a few prominent cognitive architectures with a focus on memory. 
%In a subsequent discussion we will describe how well the aforemotivated characteristics of a memory system are represented in related work.
%}

\subsubsection{Cognitivistic Architectures}
% Initial statement
In cognitivistic approaches, cognition is achieved by computations performed on internal symbolic representations. Approaches following this paradigm focus on aspects of cognition that is constant and task independent~\cite{Vernon_2021}. This way of representing information is natural, intuitive and often very performant as irrelevant data can be abstracted away.
Symbols are ideal to represent descriptive problem statements and actions. They can easily be enriched with probabilities and beliefs.
Thus, symbolic representations are usually chosen for complex high-level abilities such as planning, reasoning and language understanding.
On a semantic level, they can be used to learn from experience.
Cognitivistic agents that act in the real world require rules to derive symbols from the sub-symbolic sensory input overcoming the \emph{Signal-to-symbol-gap}~\cite{Kruger2011OACs}.

%Processing symbolic data can often be done through simple mental models, \eg if-then rules.

% EPIC
EPIC~\citep{kieras1997} (Executive-Process/Interactive Control) implements such a cognitivistic architecture. 
%It is a symbolic, rule-based approach to artificial intelligence that is designed to simulate human-like reasoning and decision-making. 
One goal of EPIC is to model cognitive executive processes with a focus on detailed timing of human perceptual, cognitive, and motor activities during multiple tasks. EPIC requires \emph{Sensory} and \emph{Perceptual Processors} to derive symbolically coded changes in sensory properties. These processors accept visual, aural and tactile inputs. The WM of EPIC is a collection of modality-specific items, but also contains perceptually unrelated information such as goals and actions, and is updated periodically. Long-term knowledge is encoded as production rules.
%EPIC also has a long-term memory module, however, there is only a one-way connection from LTM to WM, hence the LTM can only initialize the behavior of the full model. The system is not able to derive new behavior rules and put them into the LTM.

% ICARUS
Related to EPIC, ICARUS \cite{Langley2006} is a cognitivistic architecture that focuses on problem-solving.
In each recognize-act cycle, perceptual information is converted into short-term beliefs using categorization and inference and stored in the WM. Such beliefs are used to find an action sequence from the current state to a goal state.
Again, production rules to abstract from sensory input to beliefs are stored in the LTM, along with current goals and known actions.
While EPIC has a one-way connection from LTM to WM, and thus is not able to learn new production rules, ICARUS is able to learn the connection between skill execution and goal achievements in the form of new skills. Further, it is able to update skill constraints from failure executions.

\subsubsection{Emergent Architectures}
% initial statement
Symbolic representations can also have restrictions. Cognitivistic architectures that have a fixed predefined set of production rules to generate symbols from perception are less flexible and robust against a constantly changing environment~\cite{Kotseruba_Tsotsos_2017} as new production rules are not learned during operation\footnote{If some approach includes learning of new production rules from the sub-symbolic environment, it would rather be hybrid than cognitivistic.}.

Thus, another approach for flexible systems that are able to adapt and learn from the interaction with the environment focuses directly on sub-symbolic representations as emergent systems do -- typically by exploiting highly parallel models based on connectionism. Ideally, such systems evolve during operation to a full cognitive state. Emergent approaches are often heavily influenced by neurology and cognitive psychology studies, especially those of infants in which innate and learned properties and abilities are discussed and identified. The architecture itself is usually assumed to be fixed and thus belongs to the innate properties while ontogeny is achieved through learning from interaction with the environment and experience~\cite{Vernon_2007}. Because of the focus on learning from interaction with the environment, emergent approaches require a physical body~\cite{Vernon_2021}.

Unfortunately, implementations of such systems often lack transparency. It is difficult for programmers to implement inference rules or to bring in prior knowledge because this functionality is not intended by the architecture. The system should learn and adapt by itself.

% SASE
% Waren die ersten, die über self-awareness gesprochen haben im kontext von CA. Passt gut zu introspection
As an architectural approach, the Self-Aware Self-Effecting (SASE)~\cite{Weng_2002} emergent cognitive architecture is based on the idea that intelligence emerges from the interactions and connections between simple, low-level elements, rather than from a central, pre-programmed control system. In SASE, the basic building blocks of intelligence are simple, self-aware agents that can sense and act in their environment, and that are capable of forming complex networks of interactions and relationships. These agents, working together, are able to generate complex and adaptive behavior, without the need for a centralized control system.
Emergence in SASE is achieved by the combination of self-awareness, self-effectuation, and interaction. Self-awareness allows an agent to sense and represent its own state, and thus to adapt its behavior based on that state. Self-effectuation allows an agent to act upon its own state, thereby influencing its own behavior. And interaction allows agents to exchange information and coordinate their behavior, thereby creating the potential for emergent phenomena, such as cooperative and competitive behavior, and the emergence of higher-level structures and patterns.
% The Self-Aware Self-Effecting (SASE)  emergent cognitive architecture describes a computational model of the mind that is based on the principle of self-awareness and self-control. The focus on self-awareness refers to the ability of an agent to be aware of its own internal states and processes, as well as its external environment. This allows the agent to reflect on its own actions and goals, and to adjust its behavior in response to changes in its environment. Self-control on the other hand refers to the ability of an agent to regulate its own behavior and maintain its goals and objectives. It allows the agent to make decisions and take actions that are aligned with its long-term goals and values.

%MDB
%Lifelong adaption is analyzed in the MDB cognitive architecture \cite{bellas2010} through an evolutionary approach. MDB explicitly models satisfaction and motivation to achieve a goal directed behavior. 

% Deep EM
In our previous work~\cite{rothfuss2018}, we investigated how an episodic memory can be modeled based on connectionism. In this approach, the memory was tasked to convert visual experiences into a latent representation that facilitates encoding, recalling, and predicting sensorimotor experience using unsupervised learning using Variational Auto-Encoders (VAE)~\cite{VAE}.
Through the utilization of two different decoders, the learned representation can be reconstructed and predicted for the next frame.
For memory, such a learned representation is beneficial as it generalizes and compresses knowledge. In \cite{Baermann_2021}, we extend this approach to support multi-modal information, including visual information, the robot's configuration and platform pose in global coordinates, actions, plans and goal information, object labels and locations, and recognized speech as text. Using this multi-modal data, the system was tasked to answer user queries given in natural language. A deeper insight into how the memory supports this task is given in \cref{sec:casestudies:verbalization}.

\subsubsection{Hybrid Architectures}
Hybrid systems combine the best of both previous approaches. Such implementations often use symbolic representations for high-level cognitive abilities and sub-symbolic representations for learning and development. Today, such cognitive architecture is the most prevalent type~\cite{Kotseruba_Tsotsos_2017} and the most suitable for the realization of robot cognitive architectures.

% Soar
One of the earliest cognitive architectures that is still maintained and developed is Soar (State, Operator Apply Result)~\citep{Laird_1990, Laird_2012}.
The goal of the Soar project is to develop an artificial system that has similar cognitive capabilities as humans, \ie knowledge-intensive reasoning, reactive execution, hierarchical reasoning, planning, and learning from experience, and to find out what computational structures are required to support human-level agents. 

Perceptions are converted into a scene-graph-based representation and stored as such in WM along with information about targeted goals. Beyond the WM, Soar manages three different types of long-term memories: 
\begin{enumerate*}[label=(\roman*)]
    \item A procedural memory that contains skills as if-then-rules, 
    \item a semantic memory that contains facts and declarative information about tasks, and 
    \item an episodic memory that manages experiences consolidated from the WM in form of snapshots. 
\end{enumerate*}
Next to symbolic representations, Soar includes sub-symbolic processing, \ie to generate symbols from sub-symbolic percepts or to control symbolic processing~\cite{Laird2008}.
Soar has multiple learning mechanisms for different types of knowledge: Chunking and reinforcement learning acquire procedural knowledge, whereas episodic and semantic learning acquire the corresponding types of
declarative knowledge.

%\pconceptadapted{
% Act-R
%Soar shares many characteristics with other cognitive architectures, such as Act-R \cite{Laird_2022ActR, Anderson2004}. It consists of perceptual-, motor-, goal- and memory modules, \ie semantic memory and procedural memory. Perceptual modules publish data through dedicated buffers, which can be seen as a distributed working memory. All modules are coordinated through a central production system which is capable of matching, selecting and executing production rules which create a response given the information of the buffers. The goal module buffer contains information about the desired world state and helps during problem solving. Similar to Soar, Act-R is able to generalize knowledge and to derive new higher-level symbols from existing information. 
%This ability requires an understanding of the stored information, which is why we assume the knowledge to be introspective for the production system.
%}

% ISAC
The ISAC (Intelligent Soft Arm Control)~\citep{Kawamura_2008} hybrid cognitive architecture is constructed from an integrated collection of software agents and associated memories. The software agents encapsulate all aspects of perception, cognition and action and operate asynchronously. 
Perceptual events, encoded through a \emph{Sensory-Ego-Sphere (SES)}~\citep{peters2001} are first placed in the STM. An attentional network determines which events are relevant for the current situation and forwards this information to the WM. Additionally, WM temporarily holds information about motivation, goals, actions and internal processes if needed for the current task and encapsulates expectations for the future simulated by a \emph{Central Executive Agent}.
The LTM stores procedural, semantic and episodic knowledge (abstractions of SES, enriched with targeted goals, performed actions, outcomes and valuations) in multiple layers, supporting an efficient retrieval of memories. Associations are stored as state transitions in episodic memory. 
%ISAC is one of the few cognitive architectures who explicitly describe how forgetting is realized~\citep{alan2001}. 

% LIDA
LIDA (Learning Intelligent Decision Agent) was developed as a biologically inspired cognitive architecture to model all aspects of cognition in form of a global cognitive cycle~\citep{franklin2013}.
It includes a large number of cognitive modules, some of which have short-term or long-term storage capabilities.
Its cognitive cycle is divided into three phases: a perception and understanding phase, an attention phase, and an action and learning phase.
During the perception and understanding phase, sub-symbolic data from the environment is analyzed and  translated into symbols corresponding to objects, entities or events in the \emph{Perceptual Associative Memory} module. 
A \emph{Current Situational Model} holds information about an agent's present situation enriched through recall of experiences from the long-term memory modules by using local associations or similarity measures. Information about the present may decay if not stored in the long-term memory. 
During the attention phase, the content of the Current Situational Model is surveyed and the most salient information is broadcasted to all modules.
During the action and learning phase, the modules use the broadcasted information for learning and execution, \eg the \emph{Pocedural Memory Module} instantiates behaviors that can then be executed.

% CRAM
A  comprehensive system is CRAM (Cognitive Robot Abstract Machine)~\cite{beetz2010} that fuses perceptual information, semantic knowledge taken from a suite of knowledge bases, and execution results of simulated actions in order to carry out vaguely defined goal-directed everyday activities. To abstract execution plans in such tasks, CRAM uses \emph{designators} (\ie placeholders) which require runtime resolution. As soon as available, these placeholders are filled by knowledge from the internal knowledge base \emph{KnowRob2}~\citep{beetz2018}. This system contains a large-scale ontology of symbolic information used for reasoning and generalization. To integrate non-symbolic information, computable predicates are used. Episodic knowledge is represented as \emph{NEEMS} (narrative-enabled episodic memories), a first-order time interval logic expression enriched with detailed episodic low-level information, such as perceptual or procedural events and signals. Sub-symbolic information is used through
 logic interface using computational predicates, inherently grounding resulting representations and ensuring consistency with the environment. The chosen data structure thus allows inspecting information in the memory in order to use it for learning, reasoning, simulating the outcome of actions and evaluating their feasibility. Apart from gathering knowledge in the real world, episodic knowledge can also be generated or refined by simulating actions in an inner-world model consisting of a high-quality virtual reality system and physics engine. Generalization and Specialization are achieved through meta-cognitive induction. 

% MemoryX
The predecessor memory model of our work was part of a hybrid architecture \cite{Waechter2018, Vahrenkamp_2015} implemented in \armarx. 
%The \armarx cognitive architecture covers perception, action planning and execution, attention, memory, learning, and partially reasoning and prospection. 
A centralized working memory component receives arbitrary data, \ie symbolic and sub-symbolic information) stores it and eventually retrieves it when needed. The working memory is initialized via prior knowledge, a special sort of long-term knowledge containing only information provided by programmers, such as object, robot and world model, object features, and pre-learned skills.
The perception data is  passed to the WM, replacing the last instance of the same type. Thus, the WM always only holds the most recent information. Accessing the WM is done either directly or by listening to memory change events. Snapshots of situations including robot pose, robot configuration and object poses could be stored in the LTM.
Skills and actions were stored in the procedural memory in the form of symbols mapping to executable statecharts that can be instantiated and executed if needed. 
%Although the memory is able to process arbitrary types of data, those can only be inspected if the type information is known in advance. If no type information 
%Although the memory is a centralized component, running in one process on one machine, its data is accessible within the distributed system. Thus, components that produce a large amount of data create a high load on the communication channels. % have to transfer a lot of data over the network.
%In worst cases, this may block communication from other components. 
%defined before the memory during the build process. 
%This requirement makes the system very inflexible as defining new modalities requires to distribute the type information across the complete system. %change the dependencies of the whole system. 
%Associative types cannot be represented at all. %are not even possible. 
%In addition, the decision which data should be persisted in the long-term memory was made outside of the memory system, namely by components being able to directly connect to the long-term memory.
%This architecture does not contain an episodic memory. Instead, new instances of one modality always completely replaced the old ones in both working and long-term memory, making learning from experience on memory side almost impossible. 

\subsection{Comparison of Memory Systems in Artificial Cognitive Architectures}
\label{sec:cognitive_architectures:comparison}
% What are the problems of the existing architectures regarding memory?
% How to we improve the related work?
% Table about memory characteristics
% 
In artificial cognitive architectures, memory is an essential part of the system. Thus, almost every system described in the survey in \cite{Kotseruba_Tsotsos_2017} models some sort of memory. As described above, working memory, procedural memory, and semantic memory are commonly represented, while sensory memory and episodic memory are explicitly modeled by only a few.

However, existing memory systems of cognitive architectures are often criticized for having the wrong focus \cite{Wood_2012}, since in some artificial systems (\eg \cite{kieras1997, Langley2006, Vahrenkamp_2015, Waechter2018}), the memory is only assumed to be passive storage. This means that memory has no active state and that new information that enters the system has no influence on how existing information is represented, processed or interpreted.
%This is true for \eg to the working memory of SOAR or the deep episodic memory. 
Instead, memory should be an active part of the system that changes during operation. The memory itself must be able to adapt to circumstances and respond to new data. Some architectures partially implement this requiremen,\eg in Soar~\cite{Laird_1990} where the WM does not react to new data while the LTM involves learning from new data.
In addition, many cognitive architectures focus on the implementation of different types of memory without considering the interconnection between them \cite{kieras1997, Vahrenkamp_2015, Waechter2018}, associating entities of different modalities. Sometimes, only association between homogeneous data types is possible~\cite{Kawamura_2008, Laird_1990}.
%For example, the iCub and ISAC architecture implement a procedural and episodic memory but associations are only learned auto-associatively. 
Without the possibility to store multi-modal information and to associate multi-modal memories, it is difficult for the system to draw common conclusions from knowledge obtained from  different sources. Few systems hide the ability to associate information behind neural networks that learn a connection of data sources~\cite{Baermann_2021}. Such associations are however useless for applications outside the neural network, such as reasoning components.
%Thus, the cognitive system cannot learn that a specific cognitive ability, such as generating grasp candidates, is associated with an action, such as grasping.
\cite{Wood_2012}~also criticized the way how information is represented in artificial memory systems. Physiologically, all areas of memory are the same, \ie there is no structural difference between WM, episodic memory or semantic memory. Hence, we believe that memory should support an inherently episodic structure and that the information stored in memory must have a unified representation. Artificial cognitive architectures usually utilize specialized representations and manage modality-specific knowledge in modality-specific containers~\cite{kieras1997, Langley2006, franklin2013, beetz2010, Kawamura_2008, Vahrenkamp_2015, Waechter2018}.

%Finally, a memory must be able to access distributed data sources. This allows the system to be flexible and to reduce network traffic on distributed systems as the data can be stored and processed where it is produced. 
In addition to the identified requirements by~\cite{Wood_2012} and closely related to an active memory, we emphasize that introspection is another key component of memory and data representation as it allows the system to adapt its behavior based on the stored information, to learn and use the information for \eg internal simulation, augmentation and prediction. Thus, introspection is also strongly related to the ability to monitor the system's internal cognitive processes. Some cognitive architectures only focus on the perception-action coupling~\cite{Langley2006}, however, other architectures explicitly model this capability~\cite{Weng_2002, Laird_1990, beetz2010}.

In summary, we believe that a memory must be active, inherently episodic and multi-modal, associative and introspective. Only a memory that supports all these characteristics is able to adapt to new situations and respond to new data, have a unified representation of knowledge and support the overall system with data-driven services and data tracing for high-level abilities such as reasoning, explanation, prospection, augmentation or simulation. A more detailed explanation of these characteristics, including technical requirements for a memory system running on a humanoid robot system  is given in \cref{sec:requirements}.

\cref{tab:related_work} shows a comparison of the cognitive architectures described above. We omitted the architecture SASE~\cite{Weng_2002} as is it more an architectural approach, not an implemented system and thus open to many possibilities. 
The requirement that the memory adapts to new data and actively changes is fulfilled in particular by emergent and hybrid architectures. For emergent architectures, this property therefore follows directly from their definition. Hybrid architectures usually use the ability to process symbolic and sub-symbolic data for  learning and development, also fulfilling the active requirement.
Some architectures do implement an episodic memory, but they do not support an inherently episodic structure in all memory structures. On the other hand, in our previous work in \cite{Baermann_2021} we only model episodic memory omitting WM. We strongly believe that an inherently episodic structure is required for all modalities and for all memory structures.
Multi-modality is implemented by several systems. \cite{Kotseruba_Tsotsos_2017} gives a more detailed review of which modalities are supported by which cognitive architecture.
Associations are not explicitly modeled by every system. Especially associations between different modalities are beneficial as they allow the system to use correlated information during operation. To the best of our knowledge, even CRAM does not explicitly link related experiences with each other, but it does link sub-symbolic knowledge of experiences with symbolic derivations.

Finally, the ability to inspect data in the memory is naturally given for cognitivistic architectures as data is represented as symbols. Some hybrid architectures claim to have an introspective datatype as well, usually facilitating meta-cognition.

\begin{table}
\centering
\resizebox{\textwidth}{!}{
\begin{tabular}{| c | c | c | c | c | c | c | c | c |}
\setlength{\extrarowheight}{20pt}
\setlength{\tabcolsep}{12pt}
 \textbf{Cognitive Architecture} & \textbf{Paradigm} & \textbf{Active} & \textbf{Inherently Episodic} & \textbf{Multi-modal} & \textbf{Associative} & \textbf{Introspective} \\
 \hline
 \makecell{\\\cite{kieras1997}\\~}     & cognitivistic   & \xmark & \xmark & \xmark & \xmark & \cmark \\
 \hline
 \makecell{\\\cite{Langley2006}\\~}     & cognitivistic   & \xmark & \xmark & \xmark & \xmark & \cmark \\ 
 \hline
 \makecell{\\\cite{Baermann_2021}\\~}   & emergent        & \cmark & (\cmark) & \cmark & (\cmark) & \xmark \\
 \hline
 \makecell{\\\cite{Kawamura_2008}\\~}  & hybrid          & \cmark & (\cmark) & \cmark & \xmark & \xmark \\
 \hline
 \makecell{\\\cite{franklin2013}\\~}  & hybrid          & \cmark & (\cmark) & \xmark & \xmark & \xmark \\
 \hline
 %\makecell{ \\ \cite{benjamin2004} \\~}   & hybrid          & cell9 & cell5 & cell6 & cell5 & cell6 & \cmark \\\\
 %\hline
 \makecell{\\\cite{beetz2010}\\~}          & hybrid          & \cmark & (\cmark) & \cmark & (\cmark) & \cmark \\
 \hline
  \makecell{\\\cite{Vahrenkamp_2015, Waechter2018}\\~}  & hybrid          & \xmark & \xmark & \cmark & \xmark & \xmark \\
 \hline
 \makecell{\\This work\\~}             & hybrid          & \cmark & \cmark & \cmark & \cmark & \cmark \\
  \hline
\end{tabular}}
\caption{Comparison of the aforementioned cognitive architectures with respect to the identified requirements for a memory system in robotics. (\cmark) indicates that a requirement is only partially fulfilled. }
\label{tab:related_work}
\end{table}

\section{Requirements for a Memory System}
\label{sec:requirements}

%In this section, we introduce our novel memory architecture that combines experience-based learning and generative knowledge extension for our cognitive robot architecture.
This section explains the conceptual requirements already outlined and identifies the technical requirements for a memory for a complex robotic system.
%These requirements include technical aspects as such as the need of \emph{long-term} capabilities for reducing the dist space and a \emph{distributed design}.
%Afterwards, we describe how this memory system is integrated into our cognitive architecture which is implemented in the robot development environment \armarx.

\subsection{Conceptual Requirements}
\label{sec:implementation:concept}

% @Andre schau mal in das letzte Kapitel von Related Work, im Grunde gehe ich da auch nochmal über die Requirements drüber. Und in der Conclusion ist auch nochmal ne Auflistung

\begin{figure}[htb]
    \centering
    \includegraphics[width=\linewidth,bb=250 500 1150 700,clip]{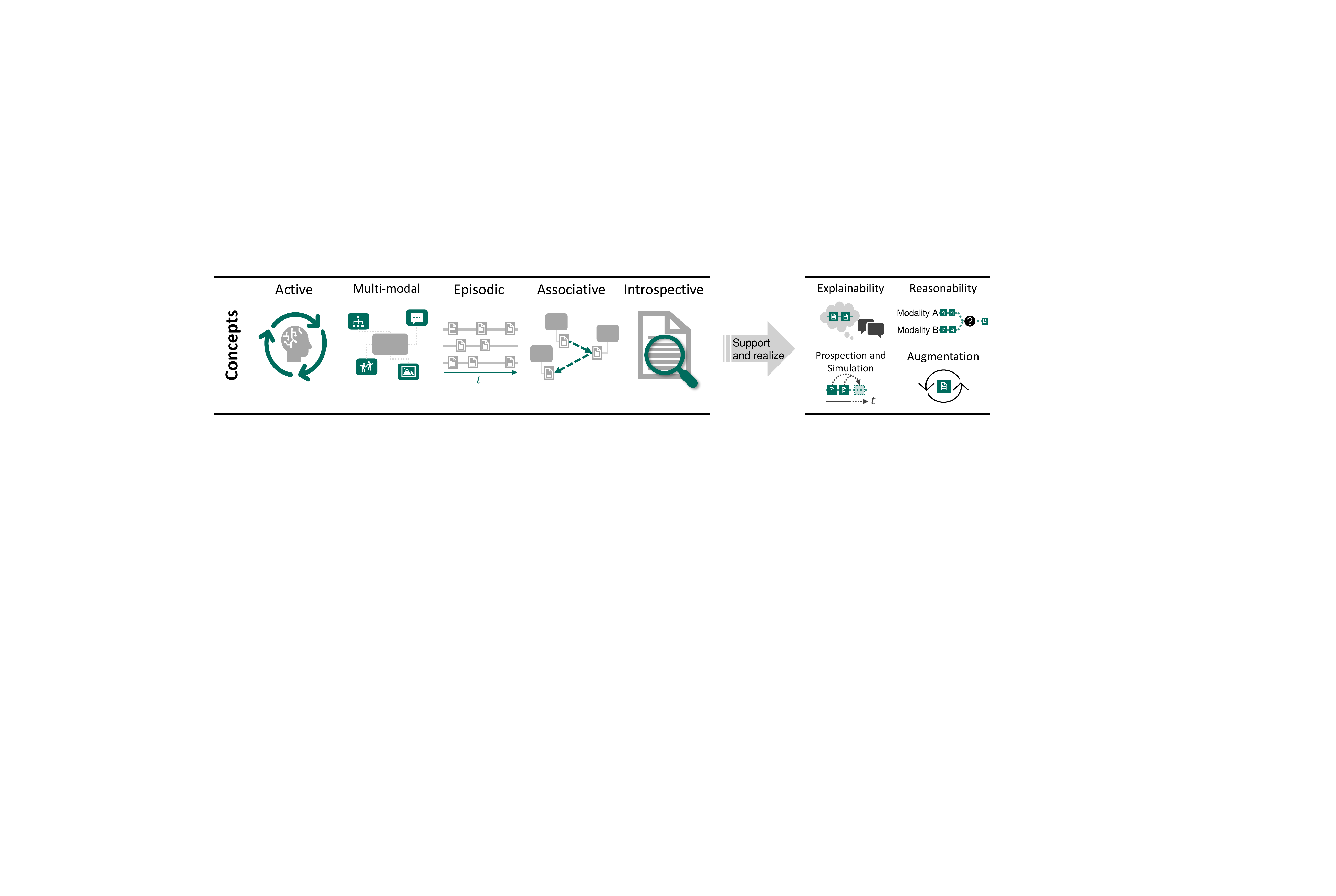}
    \caption{Identified conceptual characteristics for a memory system as  motivated in \cref{fig:motivation}.}
    \label{fig:memory_characteristics}
\end{figure}

% First draft of introduction
The memory system should centralize knowledge but also structure data flow within cognitive architectures while at the same time being efficient, scalable, flexible and understandable. 
Instead of various data exchange channels between perception, processing and execution components, general methods and interfaces should allow standardized communication between high-level and low-level.
\cref{fig:memory_characteristics} depicts required conceptual characteristics of the memory structure and the underlying knowledge representation. As described in \cref{sec:cognitive_architectures:comparison}, some memory systems of related cognitive architectures already implement those characteristics partially.

%\subsubsection{How to represent}

In the context of  humanoid robots, various sensors perceive the internal and external state. 
When performing a task such as bimanual mobile manipulation, many processing and execution components are involved simultaneously, including scene interpretation from a static and/or dynamic scene as well as the control of different parts of the humanoid robot such as \eg hands, arms and platform.
Consequently, representing \textbf{multi-modal} data in memory systems is mandatory, as is the \textbf{association} of knowledge~\citep{Wood_2012}, since most data modalities and experiences are highly correlated. In our memory, all knowledge shares a unified representation. This allows us to define general methods for processing multi-modal information. Further, we allow links between entities in the memory, thus fulfilling the requirement of associated knowledge. Those links can be established on a data level, \eg associating haptic experiences with the performed action, or on an experience level, \eg associating the task of searching an object with the interaction with the object.

%\subsubsection{What to collect}

Time-series information is hereby of particular importance.
In humans, the episodic memory is concerned with the recollection, organization and retrieval of time-series information~\citep{Tulving_1972, Tulving_1984},
\ie personally experienced events occurring at a particular time, place and context.
We believe that this is also true for semantic and procedural knowledge. 
%Such recollection of events and activities plays a key role for the acquisition of semantic knowledge in a cognitive system.
Seemingly factual knowledge may depend on the temporal context, and is only considered a fact because it changes at a very low frequency.
An intuitive example of this is how Pluto was considered a planet for the longest time, until being reclassified as a dwarf planet in 2006.
Therefore, all data, either produced periodically in streams or as a consequence of certain conditions, should be considered \textbf{episodic} and stored accordingly. Also motivated by the fact that physiologically there is no difference between episodic and working memory~\cite{Wood_2012}, we use one unified episodic structure for all parts of our memory, thus making it inherently episodic.  

%\subsubsection{How to derive}

As already stated, the memory system should not be seen as a simple, passive storage device. 
It is an \textbf{active} part of an agent's cognitive architecture~\citep{Vernon_2014, Wood_2012} which is highly influenced by the current context.
The context not only influences what we store and forget but it also influences how we store information. 
\Eg highly emotional situations will create much stronger memories which decay much slower than memories taken from a normal situation~\cite{Samsonovich_2013}. 
In particular, the memory must play a key role in learning how to derive symbols from sub-symbolic multi-modal information by allowing to learn from experience, from interaction with the environment and by trial and error.
This also means that the information stored in the memory must be \textbf{introspective}, \ie the memory must be able to analyze the information, adapt its behavior to the given data and possibly discard knowledge if that knowledge is redundant or not necessary for the current situation. In our case, for example, the memory learns predictability from incoming data. Furthermore, the memory adapts to the amount of incoming data.

%\subsubsection{?}
In order to \emph{reason} about conditions of past events, \emph{explain} why an agent acted in a particular way, to \emph{predict} how the information might change in the future or even to \emph{augment} existing or \emph{simulate} new experiences we conclude that our memory system has to be \textbf{multi-modal}, \textbf{associative}, \textbf{episodic}, \textbf{active} and \textbf{introspective}.

%\subsection{From Challenges to Requirements to Paradigms}
\subsection{Technical Requirements}
\label{sec:implementation:requirements}

\begin{figure}[htb] 
    \centering
    \includegraphics[width=\linewidth,bb=180 450 800 650,clip]{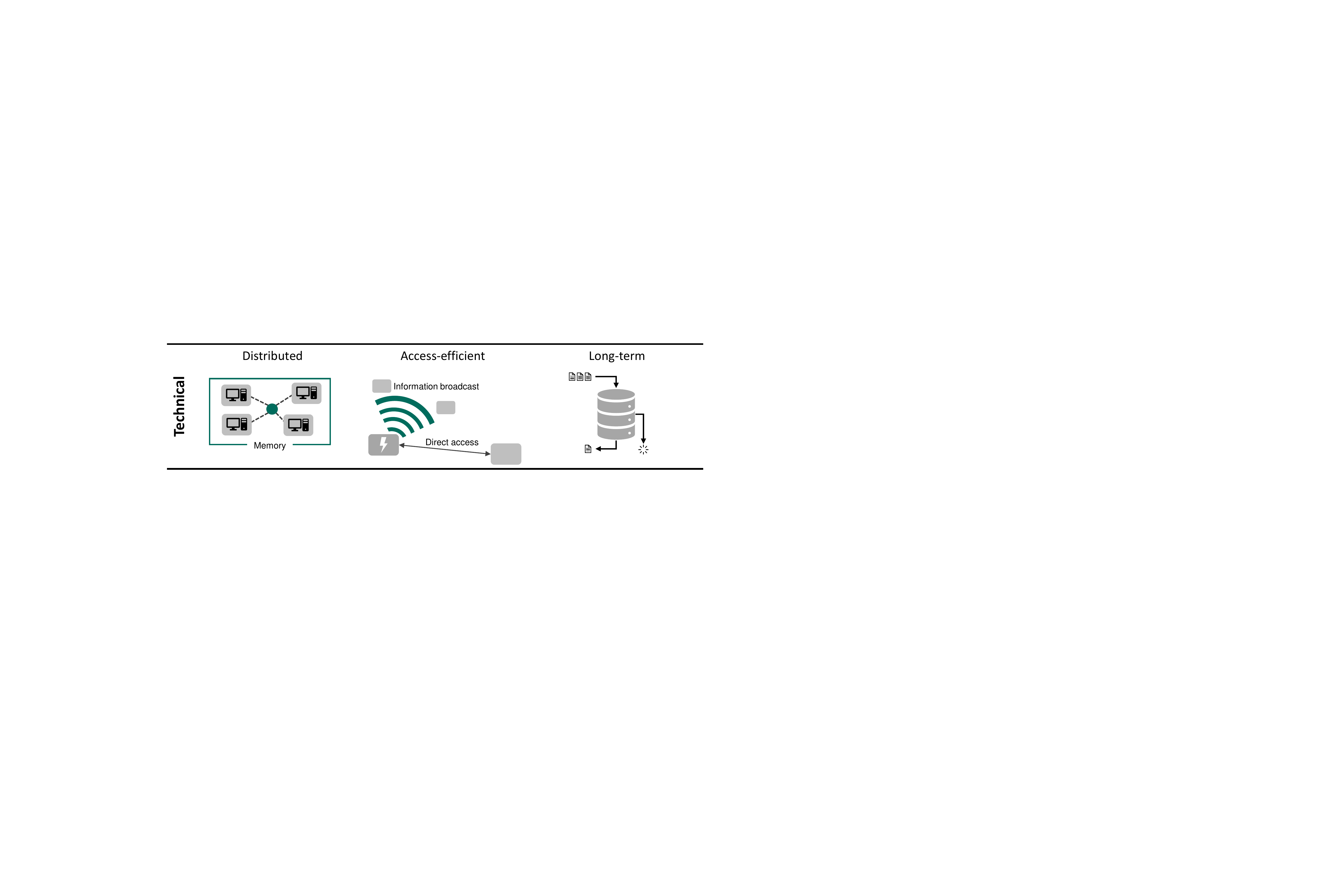}
    \caption{Technical characteristics of a memory system as part of a robot cognitive control architecture.} %This figure should be seen as an extension to \cref{fig:memory_characteristics}.}
    \label{fig:memory_characteristics2}
\end{figure}

Complex  humanoid robotic systems further pose several technical challenges regarding the implementation of a memory system of their cognitive  architecture. 
In this section, we highlight these challenges, identify the resulting technical requirements, and derive necessary paradigms. The requirements  summarized in \cref{fig:memory_characteristics2} provide answers to the questions of (i) where to run the memory system, (ii) how to improve the efficiency of access, and (iii) how to deal with space limitations.

\subsubsection{Where to Run the Memory}

Complex robotic systems such as humanoid robots are data-intensive systems that use several special-purpose computers connected to each other via common interface (usually Ethernet) to process sensory data and control the action execution. 
%It is hard to transfer the masses of data produced by the robotic system through such interfaces.
To reduce the throughput and response times, in particular in the case of visual task,  a cognitive memory architecture must be designed in a \textbf{distributed} manner, with 
%special purpose 
memories located directly on the machine where the data is produced.
This makes extending the  system at run-time easier, as additional 
%special purpose 
memories can be enabled/added on demand, or disabled to save resources when not needed anymore.

\subsubsection{How to Improve Efficiency of Access}
% Wie mit unterschiedlichen Arten von Datenquellen umgehen
% efficient memory access paradigm
% in eval erwähnen
Another possibility to reduce the throughput is to adapt to the production frequency of the data source.
In general, we can distinguish between periodically produced and event-driven generated data, \ie the data is either produced periodically in streams, or as a consequence of certain conditions, thus being event-driven.
For systems with limited communication bandwidth, it is desirable to avoid unnecessary requests to the memory, for example, if no new data is yet available. This is especially true for event-driven data sources, where data may be produced at irregular intervals. For this reason, the memory is able should be able to inform clients when new data is available. Clients can then decide whether to request the memory specifically or to wait for the next update, making the memory \textbf{access-efficient}.

%Both ways have in common, that they are \textbf{episodic}.
%We think that this is an important insight, because even (seemingly) factual knowledge is dependent on the temporal context or the situation, and it is only considered a fact because it changes at a very low frequency.
%An intuitive example of this is how Pluto was considered a planet for the longest time, until being reclassified as dwarf planet in 2006.
%Thus, every bit of information must be tied to the exact point in time it was produced in.

\subsubsection{Assessing What to Store}

Humanoid robots are equipped with a variety of sensors that generate large amounts of data, potentially at high frequencies.
A memory architecture thus needs to be able to handle such large data volumes, providing \textbf{long-term} capabilities due to limited main storage.
Since the storage system of a humanoid robot cannot be large and fast enough to simply store everything, the data must be assessed and reduced according to their importance.
This can be done by primarily storing data that was produced at or near significant points in time, \eg keypoints, identified by higher-level cognitive processes, and/or using dimensionality reduction approaches to construct meaningful latent spaces. 
%, for example by employing autoencoders, which would further also allow faster access and comparisons between data points

Depending on the use case, specific models can also be used to aggregate data into a dedicated representation.
Finally, the  system must also be able to assess and delete data that has already been stored if it is outdated, irrelevant, or proves to be incorrect.
Overall, in terms of assessment, we need to store data that allows better execution of the robots actions, analysis and reasoning on recorded episodes, and the prediction of future states of the robot itself or its environment. 

Note that this requirement is strongly linked to \emph{long-term memory} and \emph{forgetting} in biological systems. However, since we look at memory from an engineering perspective, we evaluate this property to be technical. If a system with unbound memory and unbound computational resources were available, then long-term storage and forgetting may not be necessary. The conceptual requirements of a memory motivated in \ref{sec:introduction} apply to long-term memory as well.
With this, we conclude that, in addition to the motivated conceptual requirements, our memory system needs to be \textbf{distributed}, \textbf{access-efficient}, and \textbf{long-term}. 
%In the following section, we will explain how we implemented a memory system building upon these paradigms.

\section{The Memory System}
\label{sec:implementation}
\label{sec:implementation:system}

\newcommand{\memid}[1]{{\textsl{#1}}}

In the following, we  describe how we integrated the identified requirements in a memory system for our cognitive control architecture shown in \cref{fig:armarx:architecture}. Hereby, we rely on a multi-modal data representation to support introspection and fulfill key requirements of a suitable knowledge representation for robot cognitive architectures~\cite{Paulius_2019}.
Similar to its predecessor described in~\cite{Vahrenkamp_2015, Waechter2018}, our novel memory framework consists of:
\begin{enumerate}
\item \textit{Sensory Memory (SM)} where data is held for a very short duration until it may be passed to the working memory.
\item \textit{Working Memory (WM)} that holds the current state of the world and the robot’s internal state in a volatile storage. It can be updated by the SM through perceptual processes or by cognitive processes, \eg  recalling an episode from the Long-term Memory or prior knowledge. 
\item \textit{Long-Term Memory (LTM)}, which complements the WM, provides persistent storage capabilities and encodes the information into a more graduated representation. 
%This representation focuses on the generalization of the data, as it allows multiple data points to be summarized. Nevertheless, the representation should still support introspection and thus predictive ability. %In order to minimize the used amount of disk space, the LTM makes use of different encoding and augmenting strategies and it offers the possibility to improve these strategies from the incoming data. Further, it provides a general tool for data recording.
\item \textit{Prior Knowledge (PK)} contains information that is provided by the programmer and thus already known to the robot. During startup, the WM is initialized with this known data such as robot, objects and environment models, pre-defined motion trajectories, etc.
\end{enumerate}

First, we will describe the working memory, which is the part most clients are directly interacting with. Afterwards, we present the long-term memory and its learning capabilities and which adopts the working memory's principle structure but provides a more permanent storage than the working memory. 
%TODO: Sensory memory and PK are not interesting here

From a simplified perspective, the memory system can be viewed on three levels of detail:
\begin{enumerate*}[label=(\roman*)]
    \item The \emph{distributed memory system}, which is a collection of memory servers running in their own processes, 
    \item a single \emph{memory server}, which stores data in episodically structured segments, and 
    \item a  single \emph{data instance}, which holds data in a general, interpretable format.
\end{enumerate*}

\subsection{Distributed Memory System}

The \armarx memory system is a distributed system implemented through several \emph{memory servers}. 
Each memory server is a separate \armarx component, \ie a process communicating with and providing interfaces to other components via a middleware.
All memory servers offer a common interface for, among others, reading and writing data. A concrete memory server may provide specialized interfaces for its respective modalities (\eg objects). However, we want to emphasize that all memory servers have the same structure. All memory servers are able to hold arbitrary data -- no matter if it's symbolic or sub-symbolic information. Thus, the memory servers are not modality specific. 
All memory servers register themselves in the \emph{Memory Name System} (MNS) on startup. 
The MNS is a central registry, %component (but not a memory server)
which allows memory clients to get access to 
the memory server handling a specific modality given a human-readable ID of the modality (\eg ``Object/Instance'').
This is similar to and inspired by the Domain Name System (DNS) of the internet, which resolves human-readable URLs to machine-readable IP addresses. %
Memory servers can be added at any time and distributed to several computers, distributing the load and reducing the network traffic and response times by running memory servers close to related hardware and memory clients such as a camera and our visual perception pipeline, greatly supporting the system's scalability. 

% Architecture
\begin{figure}[tb]
    \centering
    \includegraphics[width=\linewidth, bb=0 0 1500 1000,clip]{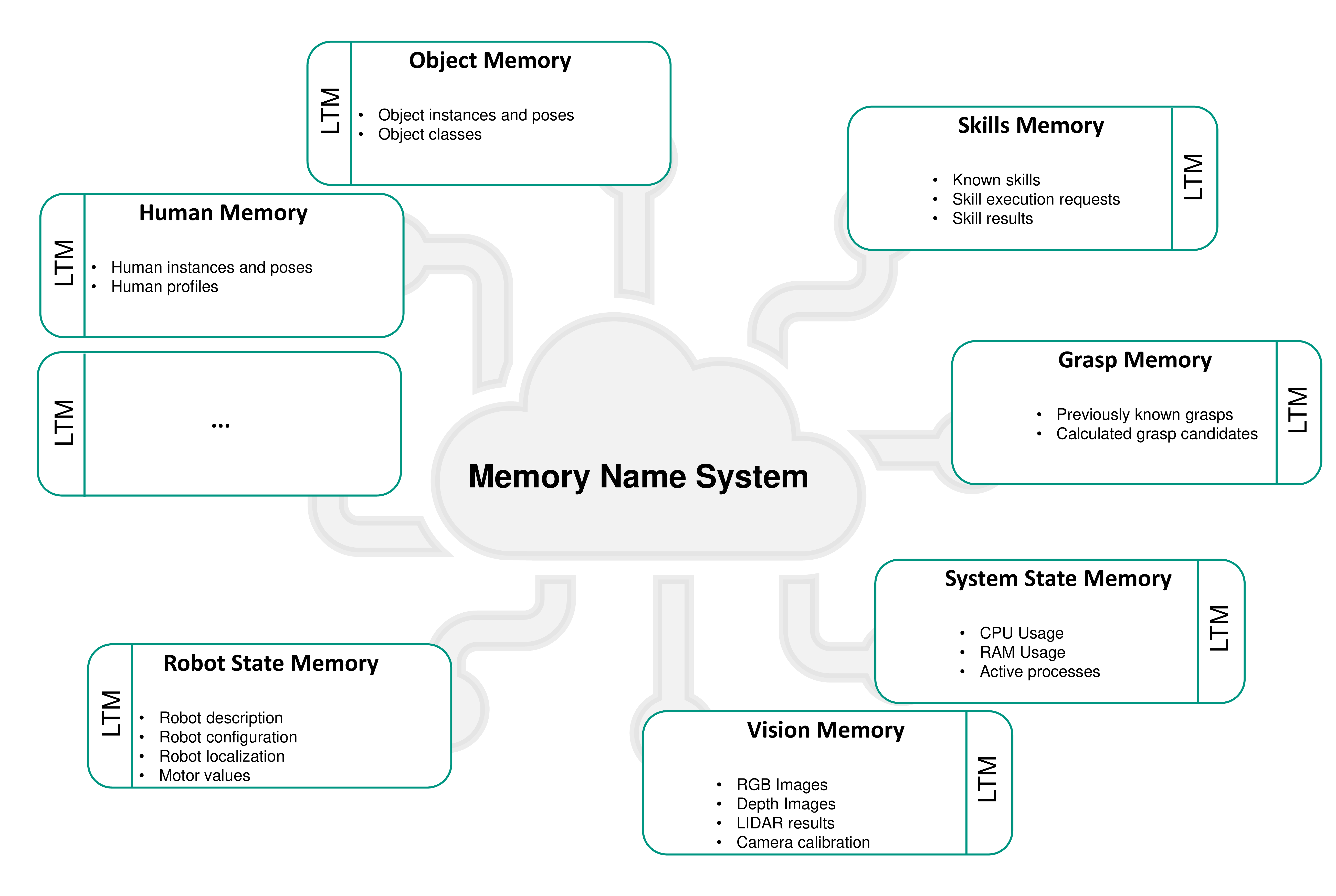}
    \caption{The distributed memory system with some of the memory servers running in \armarx. Each memory is running in a separate process and may run on a different machines. All memories are connected through the MNS. Each memory server manages its own LTM.}
    \label{fig:system_overview}
\end{figure}

As shown in Figure \ref{fig:system_overview}, the distributed memory system can be visualized as a cloud of memory servers, running in separate processes on different machines, each one extending the system with new modalities. The robots' full memory is formed by the union of all distributed memory servers. Each memory server manages its own working memory and a corresponding persistent storage. 
Working memory and long-term memory of a single memory server share the same structure, \ie hierarchical segments with a temporal structure with data in a general and introspective format.  The working memory part is limited in space while the long-term memory part is not.
The working memory can be accessed by clients to read and write data in a hierarchical structure. 
By querying the memory servers using temporal cues, the user can get access to the stored information. 
The memory servers accept precise queries to return data for a precise point in time as well as broad queries to return approximate information. 
Further, the working memory can notify clients as soon as they receive new information thus fulfilling the requirement of access-efficiency. The memory servers are used to distribute and structure the data -- usually, each memory server holds information related to a specific modality, sensor or functionality to make the system more understandable for users. \Eg the \memid{Object} memory holds all information related to objects such as object class information (name, parent classes, meshes) or concrete object instances (instance name, pose).

%\iffalse
\begin{table}
    \setlength{\extrarowheight}{3pt}
    \newcommand{\tablerowbase}[7][\\]{%
        % #2      % No
        % & 
        #3    % Level
        % & #4    % Key
        % & \makecell{#5}    % Desc
        & #5    % Desc
        & #6    % Example  
        & #7    % Example Key
        #1      % \\ 
    }
    \newcommand{\tablehead}[7][\\]{%
        \tablerowbase[#1]{\bfseries #2}{\bfseries #3}{\bfseries #4}{\bfseries #5}{\bfseries #6}{\bfseries #7}
    }
    \newcommand{\tablerow}[7][\\]{%
        \tablerowbase[#1]{#2}{\bfseries \makecell[lt]{#3}}{#4}{#5}{#6}{\small \memid{\makecell[lt]{#7}}}
    }
    \small
    \centering
    \begin{tabularx}{\linewidth}{lXXl}
        \toprule
        \tablehead{\#}{Level}{Key}{Description}{Example}{Key} 
        \midrule
        \tablerow{1}{Memory}{Name}{Collection of semantically related modalities}{Data related to Grasp planning and execution}{Grasp}
        \tablerow{2}{Core \\ Segment}{Name}{Homogeneous container of a specific modality.}{Grasp affordances}{Affordance}
        \tablerow{3}{Provider \\ Segment}{Name}{Sub-segment containing data of a single provider}{Results of a grasp planner}{MyGraspPlanner}
        \tablerow{4}{Entity}{Name}{Physical thing or concept evolving over time}{Grasp affordances of a specific object}{blue-cup}
        \tablerow{5}{Entity \\ Snapshot}{Time Stamp}{State of entity at a specific point in time}{Grasp affordances at time $t$}{2022-02-18 \\ 13:06:56.492182}
        \tablerow{6}{Entity \\ Instance}{Index}{One data instance at a point in time}{Second grasp affordance}{1}
        \bottomrule
    \end{tabularx}
    \caption{Levels of the working memory data structure.}
    \label{tab:memory-structure}
\end{table}
%\fi

\begin{figure}[tb]
    \centering
    \includegraphics[width=\linewidth,bb=0 250 500 400,clip]{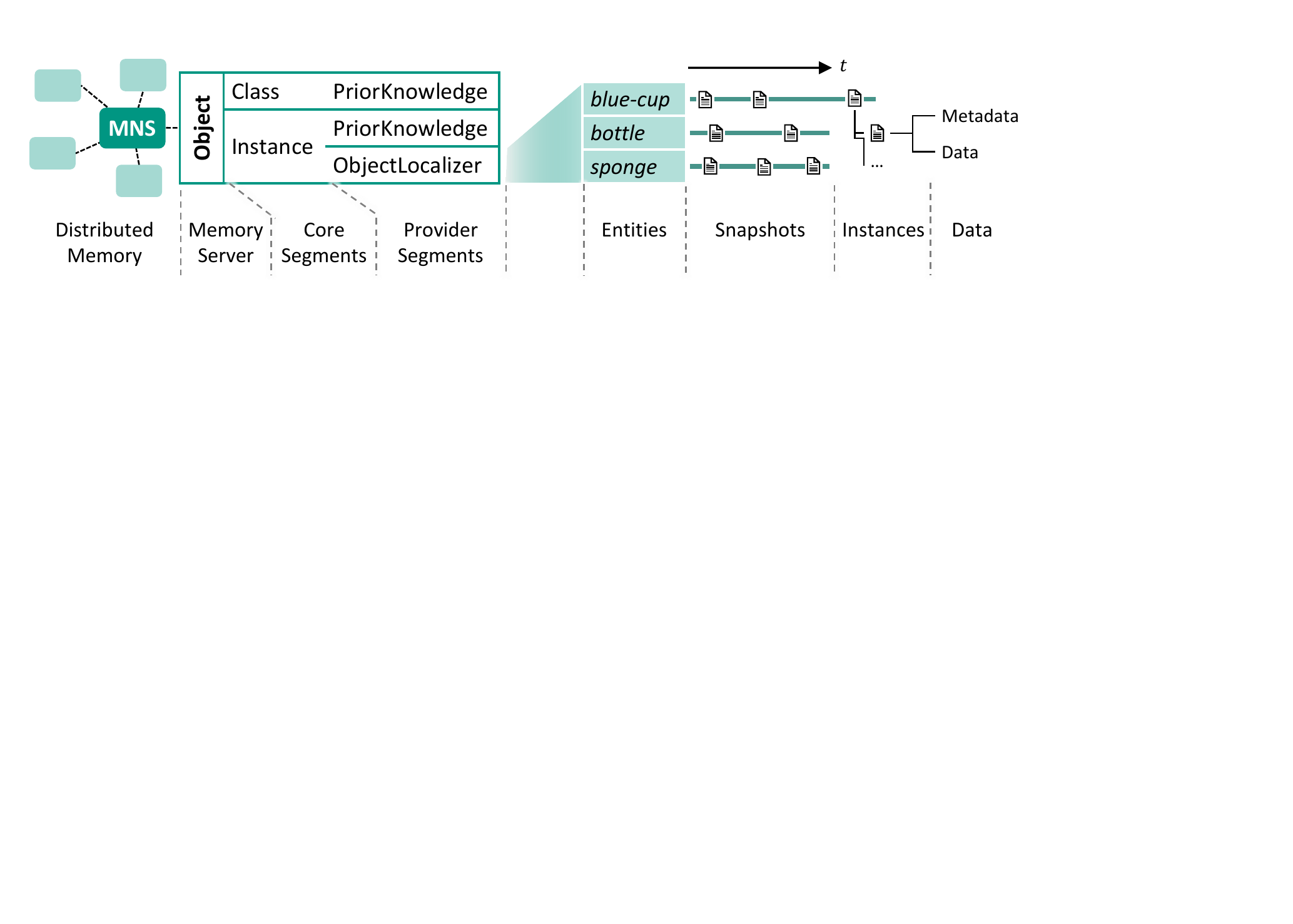}
    \caption{
        Hierarchical memory structure.
        The distributed memory system comprises several memory servers which are registered in the memory name system (MNS).
        A memory server has a name (\eg \memid{Object}) and multiple core segments (\eg \memid{Class}, \memid{Instance}) specifying the stored modalities. 
        Each core segment can have multiple provider segments (\eg \memid{PriorKnowledge}, \memid{ObjectLocalizer}).
        A provider segments stores entities (\eg \memid{blue-cup}, \memid{bottle}), which represent timelines of a thing or concept evolving over time.
        Each observation or update of an entity creates a snapshot at a specific point in time.
        Each snapshot can store multiple instances, which finally contain the payload data.
    }
    \label{fig:memory-structure}
\end{figure}

%\cref{fig:system_overview} lists more examples.

To free up space in the WM, information may be moved to the LTM. 
The decision is based on the holding duration, space limitations or internal statistics whereby associations are considered as well. If entity $A$ has not been queried recently, but is associated to entity $B$ that is accessed very frequently, $A$ will not be moved to the LTM. 
Another alternative is to make that decision depending on the system state. A special \emph{System State} memory server can be used to watch the CPU and RAM usage and to change the behavior, \eg to reduce or increase the maximum size of all memory servers running on this machine. 
%
%If data exceeds a minimum holding period without getting queried or if the amount of data in the working memory reaches a limit, information is moved to the long-term memory. 
%Such conditions must not be fixed. A special \emph{System State Memory} can be used to watch the CPU and RAM usage and to change the behavior of the memory servers. 
%
Both, the working memory and the long-term memory hold the stored information episodically, \ie they hold a dictionary mapping from timestamps to data instances. Additional meta-information such as the name of the provider or the time it took to transfer it to the memory can be used to analyze the data.  

The memory servers implement the aforementioned concepts of a distributed memory system and the inherently required episodic and associative structure. Currently, our working memory holds data in plain text. There is no encoding as most components require precise information about how the robot is moving or where objects are located.

%They can communicate with each other directly or indirectly and relate to each other via references to other memory servers.

% TODO: meta information such as provider info, time provided, confidence etc...

%\subsubsection{Memory Servers}

\subsection{Hierarchical Data Structure}
\label{sec:implementation:system:structure}

The general working memory data structure comprises multiple levels listed in \cref{tab:memory-structure} and illustrated in \cref{fig:memory-structure}, starting at a memory server down to a single data instance.

Data structures of level \emph{Memory} as described in \cref{tab:memory-structure} is a semantic collection of one or many modalities, such as objects, actions, skills, images, locations, relations, the robot itself, and so on. 
This memory is usually represented by a single memory server and is identified by a memory name, \eg \memid{Object} or \memid{Navigation}. 

\emph{Segments} are homogeneous containers of specific modalities, for example, object classes, object instances, grasp affordances, grasp actions, images, point clouds etc.
The data structure contains two levels of segments: \emph{core segments} and \emph{provider segments}.
Core segments are usually defined statically by the containing memory and determine a core data type shared by all data in this segment. Incoming data not fulfilling this data type will be reinterpreted as such type if possible. 
Provider segments are created by clients writing to the memory, \ie providing data, at run-time. 
They serve three purposes: 
(1) They identify the source of the data, 
(2) they create a separate namespace for the provider to avoid conflicting names on the next level, and
(3) providers are allowed to extend the core segment's data type, such as pixel-wise labels in addition to bounding boxes in image segmentation, allowing some specialization while still being compatible with the general representation.
% All core segments and provider segments have unique names inside their parent element, so that the example provider segment in \cref{tab:memory-structure} can be identified by the ID \memid{Grasp/Affordance/MyGraspPlanner}.

The elements of a provider segment are \emph{entities}.
An entity is a physical thing or concept that exists and evolves over time.
Examples are images captured by a camera or object instances reported by an object pose estimation component.
Entities are identified by names inside their parent provider segment.
As entities are supposed to evolve over time, we represent an entity as a timeline or history states over time.
Therefore, entities and thus the whole memory are inherently episodic, fulfilling the aforementioned requirement of an inherently episodic memory.
These discrete states are called \emph{entity snapshots}.
A snapshot can represent a new sensory observation, the result of a cognitive process, or in general an update to the knowledge of the robot.
Finally, each snapshot contains a list of an arbitrary (and potentially varying) number of \emph{entity instances}. 
In practice, some modalities often only have one instance per snapshot, such as object instances and the robot's state, 
but others usually store multiple instances, \eg two images of a stereo camera or multiple extracted grasp affordances of an object instance.

All elements of the hierarchical memory structure have keys, which are unique inside their parent container.
The keys of memories, core segments, provider segments, and entities, are textual names.
The keys of entity snapshots are time stamps, while a single instance has an integer index.
This way, all elements on all levels can be identified by a unique memory ID, which is the path through the hierarchy down to the respective element.
For example, the entity instance in \cref{tab:memory-structure} is identified by \memid{Grasp/Affordance/MyGraspPlanner/blue-cup/2022-02-18 13:06:56.492182/1}, 
while the mere entity is identified by \memid{Grasp/Affordance/ MyGraspPlanner/blue-cup}, 
and the ID \memid{Grasp/Affordance} refers to the core segment for grasp affordances. 

These memory IDs allow to create associations and cross-references between instances of different modalities, fulfilling the requirement of an associative memory.
% For example, a localized object instance can be associated to its static object class (an entity), 
%and a .
Ideally, the output of a processing steps references the respective input. Thus, complete traces of data through the processing pipeline are stored in the memory, which could enable a cognitive agent to reason about the whole process leading to, for example, a failed grasp attempt. Additionally, these traces help programmers to debug and improve processing pipelines.

\subsection{Reading and Writing: Queries, Commits and Updates}

\label{sec:implementation:system:rw}

Among others, memory servers offer interfaces for reading data from and writing data to the working memory.
% A memory server may be read-only, \ie not offer the writing interface, which is useful for sensory memories
% which receive their data from a hardware device.
Writing is done via \emph{commits}. 
A commit is a bundle of \emph{entity updates} which are written to the working memory
% in a single network call
.
Each entity update adds or updates one snapshot to an entity.
%(if the entity does not exist yet, it is created by the server). 
Therefore, to create an entity update a client must specify
(1)~the entity's ID,
(2)~the snapshot's timestamp, and
(3)~the data of each instance at this point in time.

\begin{figure}[tb]
    \centering
    \includegraphics[width=\linewidth, bb=0 350 500 450,clip]{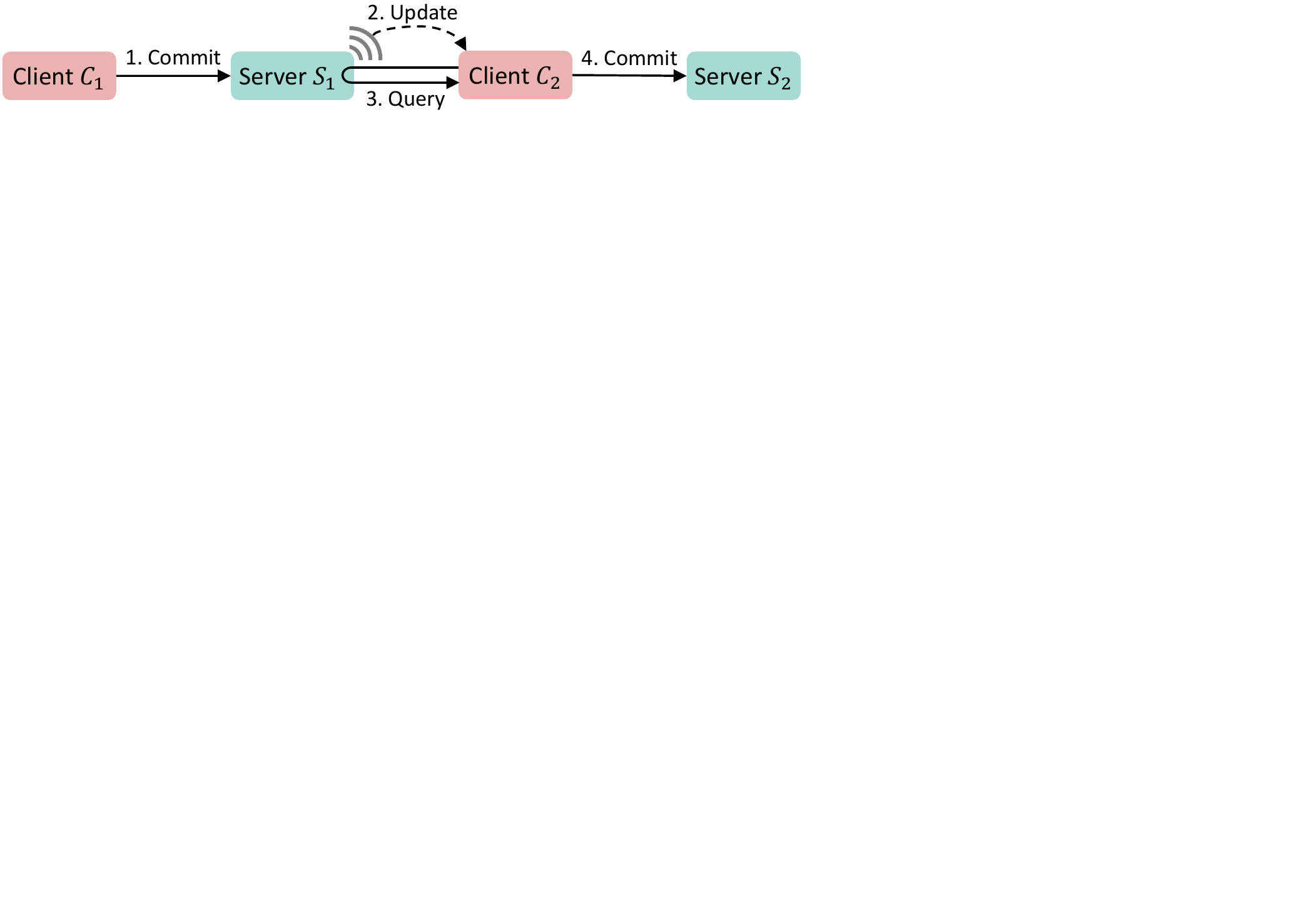}
    \caption{
        Event-driven pipeline scheme.
        (1) Client~$C_1$ writes data to server~$S_1$ via a commit.
        (2) Server~$S_1$ broadcasts an update notification, which is received by client~$C_2$.
        (3) Client~$C_2$ performs a query to~$S_1$ to read the updated data.
        (4) Client~$C_2$ processes the data and commits the result to another server~$S_2$.
    }
    \label{fig:event-driven-pipeline}
\end{figure}

Each time a memory server receives a commit, it updates its internal data structure and broadcasts a notification message to listening clients. 
This message contains the IDs of the memory snapshots that have been updated by the commit. 
% All clients and other memory servers may subscribe to this broadcast. 
When a client receives a notification, it can decide whether it needs to react to the updates; in this case, it can read the updated data from the server and process it. 
These update notifications allow building data processing pipelines in an event-driven manner as can be seen in \cref{fig:event-driven-pipeline}.

To read from a memory server, a client performs a \emph{query}.
The structure of a query follows the memory data structure: 
For each level of the hierarchy, it specifies which elements should be part of the query result based on their keys.
For each of the name-keyed levels (memory, segments and entity), the client can request 
all entries, a single entry with a specific name, or all elements whose name matches a given regex.
For the entity snapshot level, there are many options based on time stamps and logical indices. 
Commonly used snapshot queries include
the single latest snapshot (the current state), 
the snapshot at a specific point in time, 
the snapshots in a time window (\eg in the last \SI{10}{\second}),
and
the latest $n$ snapshots. % (\eg to show the latest history in a user interface).

Multiple queries at the same level can be combined to create a conjunction, \eg to request data from two specified core segments.
In this case, each core segment query has its own sub-queries for the lower levels.
As a result, queries have the same tree-like structure as the memory data structure, 
and can be flexibly combined to create complex queries.

\subsection{Interpretable Data Format}
\label{sec:implementation:system:idf}

\newcommand{\dataobject}[0]{data-object}
\newcommand{\dataobjects}[0]{data-objects}

\newcommand{\typeobject}[0]{type-object}
\newcommand{\typeobjects}[0]{type-objects}

The ability to inspect and analyze data is by default not supported by the used programming language \cpp. \cpp is standard in the context of robotics frameworks in terms of performance. 
Therefore, we introduce the \armarx \emph{Interpretable Data Format} (IDF), a special format that allows memory servers to use a common introspectable data representation. 
In its very essence, this data format implements a recursive variant formulation with special extensions that are frequently used in robotics (such as matrices, orientations, images or point clouds) and minor optimizations for network transfer.
All data in the robot's full memory is a composition of basic \dataobjects{} (int, long, float, double, bool, string and multi-dimensional byte array) and recursive \dataobjects{} (list and dict). These objects can be inspected at run-time, are performant, are transferable over the network, and offer a strong interface via standard \cpp classes.

Segments of our memory can be annotated with static type information
%, that is again a recursive variant based formulation
. 
That is a mirrored representation consisting of basic \typeobjects{} (\ie int, long, float, double, bool, string, time), special \typeobjects{} which get encoded into multi-dimensional byte arrays (\ie matrix, orientation, image, point cloud) or container \typeobjects{} (\ie list, tuple, dictionary, object, pair). Type annotations do not contain data and only provide additive information for the underlying data objects. 
%\Eg the data object of a list contains a sequence of all elements of that list. These elements are variant data objects. The type object for that list only has one member representing the accepted type of that list.
%\cref{fig:aron_mapping} shows the contents of the data objects and the type objects as well as their relation. 
Thus, the relation of \dataobjects{} and \typeobjects{} is not bijective.
Type annotations are optional. Even if some segment has no knowledge about the detailed type, it knows which members are present, their name and it can inspect the data. 
This property is useful when implementing type-agnostic procedures which are applied to all data objects in the memory, no matter what they represent.

%\Eg for data objects with only numerical values it is possible for the memory to provide predictions using linear regression over the last $n$ entries in the episodic storage of the memory server. However, if type information is present, the memory server can use that specialized knowledge (e.g. knowledge about special regression models, min or max constraints of single members, ...) to provide better predictions.

Users can specify the static type information through \emph{XML}. Our system uses code generation as an abstraction mechanism to separate the type description from the implementation. 
Given the type information in XML, IDF automatically generates so called \emph{business objects} for a particular implementation language. It also generates common interfaces to cast IDF data objects to its corresponding business object and vice versa.
These business objects are handier to use for clients as the auto-generated \cpp classes allow programmers to make use of all programming language features, static type checking and code completion. 
%These business objects are used by the clients to interact with the memory, but they can also be used as a data representation for peer-to-peer communication.

Separating type and data makes data objects interchangeable. Even if a data object only fulfills a part of some type, it can be casted to the corresponding business object. Unmatched members stay uninitialized.

Another benefit of using code generation mechanisms is that our system is able to generate similar code in different languages while maintaining the network transfer ability of our middleware. Thus, memory servers and memory clients are not required to be implemented in the same language.
Currently, IDF supports \cpp and \python as target implementation languages. 
%For being able to transfer objects via the network, IDF has a mirrored representation in our data transfer middleware to which IDF is automatically converted when required. 
%Further, the generated business objects automatically cover the conversion from and to IDF, human readable formats and compressed formats. 
%During conversion, these auto-generated methods also keep track of asserting that all specified constraints are fulfilled. Through the code generation, users usually do not have to know about this specialized introspectable format as they can use the business objects and all conversion is done behind the scenes.

% TODO: besserer übergang to LTM

\subsection{Long-term Memory and Deep Episodic Memory}

\label{sec:implementation:system:ltm}

While the working memory parts of memory servers represent a volatile in-place memory that is limited in space, the long-term memory part can hold a large amount of data for a longer period. The exact hold time depends, for example, on the access frequency and importance of the data and on the parameterization of the memory server. In our implementation, the long-term memory will be stored on the hard drive of the machine on which the memory server is running. This reduces the network traffic and increases the response times of the system as data is not transferred via the network to a central storage. 
% Through unified interfaces and online IDF conversion tools, information can be stored in SQL databases, NoSQL databases or simply on the hard drive (\eg in JSON or HDF5 \cite{HDF5} files). 
To be space-efficient, the long-term memory converts memory instances into a compressed but still generative and predictable format. 
%This offline conversion is learned unsupervised through the utilization of auto-encoders. 
Further, the long-term memory is able to filter incoming and to forget already stored information.

\begin{figure}[tb]
    \centering
    \includegraphics[width=\linewidth,bb=0 0 780 550,clip]{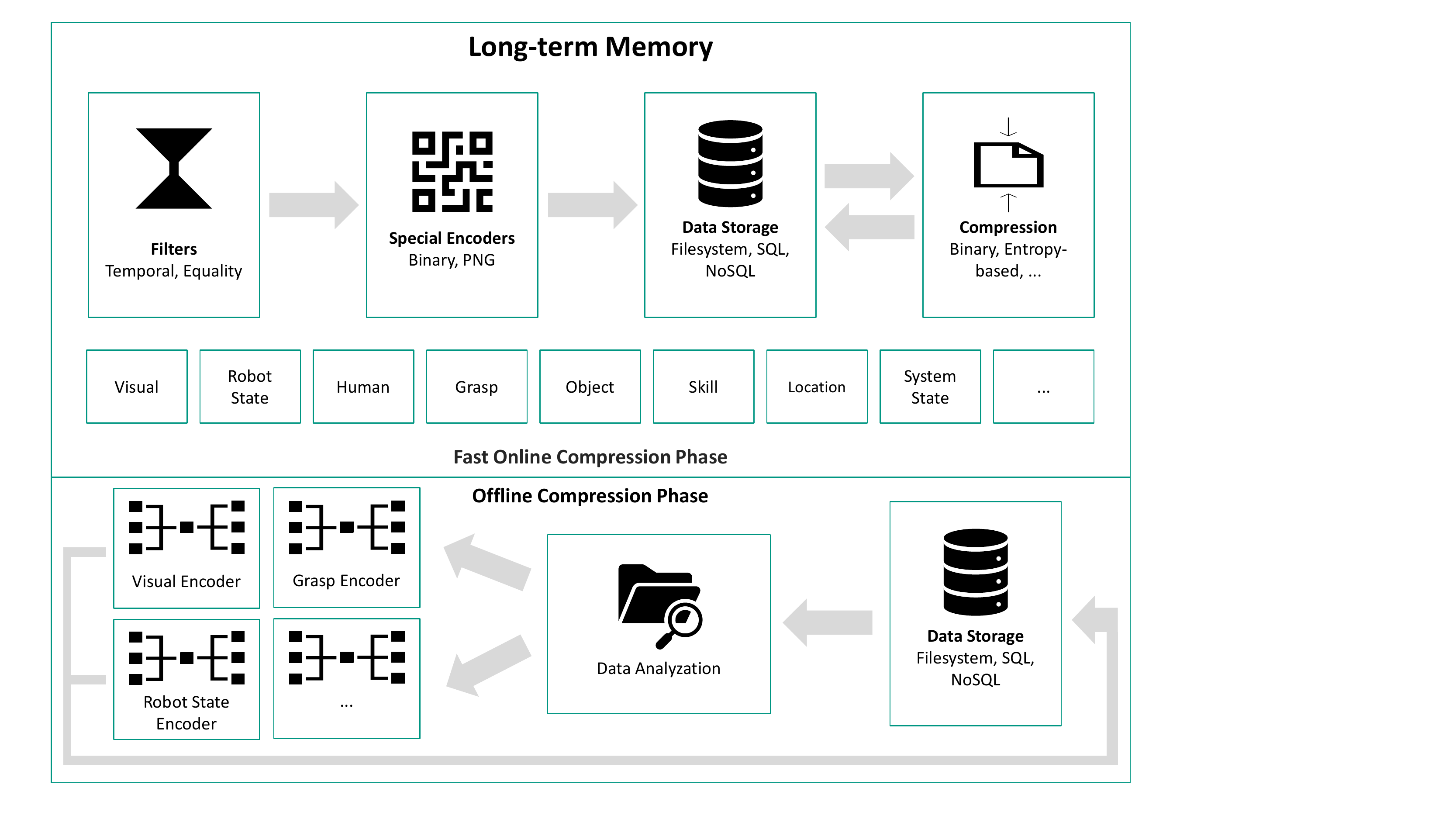}
    \caption{The full pipeline to filter, store and compress data in the long-term memory part of 
    the memory servers.}
    \label{fig:ltm:pipeline}
\end{figure}

The full pipeline to filter, store and encode data is shown in~\cref{fig:ltm:pipeline}. In an online phase, data is filtered based on their frequency or using fast similarity measures compared to other entities of the same type. Remaining data gets compressed using standard compression methods (\eg JPEG~\cite{JPEG} or MPEG-4~\cite{MP4} for images and videos, ZIP~\cite{Huffman_2006} for arbitrary snapshots) in order to reduce the final file size. Which online compression technique is suitable depends on the parameterization of the memory server. Users can choose to not filter incoming information and to only use lossless compression, which allows the long-term memory to be used as a general tool for data recording.

In an offline phase, \ie when the robot is inactive, the memory further compresses and abstracts the episodic information using machine learning techniques into a \emph{deep episodic memory}. We utilize Auto-Encoders (AE) to convert the snapshots into a generalized but still predictable format, similar to \cite{rothfuss2018, Baermann_2021}. The ability to encode information evolves and improves over time as more training data is available.

Currently, the encoding works for data types that do not contain strings or lists with a varying number of elements as they can not be converted into a well-defined input format for machine learning. Our framework automatically extracts those kinds of information and stores them separately, while the rest of the information is passed to the AE.
This representation does not affect the efficiency of existence checks of snapshots. If a client queries a snapshot that has already been moved to the LTM and converted into its latent representation, the memory can still access corresponding meta-data. If the memory finds a matching snapshot, it will decode the information, convert it back to IDF and finally return it to the WM. The WM can then forward the result to the client. 
Thus, at run-time, each long-term memory part of the memory servers manages its data in two representations: (1) recent information, that is filtered and compressed using the online methods and (2) older information, that has already been brought into a latent representation. 
%We implemented the autonomous learning framework with tensor flow. The code is available on GitLab\footnote{\url{A link will be provided once the paper got accepted}}.

\begin{figure}[ht]
    \centering
    \includegraphics[width=\textwidth,bb=0 0 1800 300,clip]{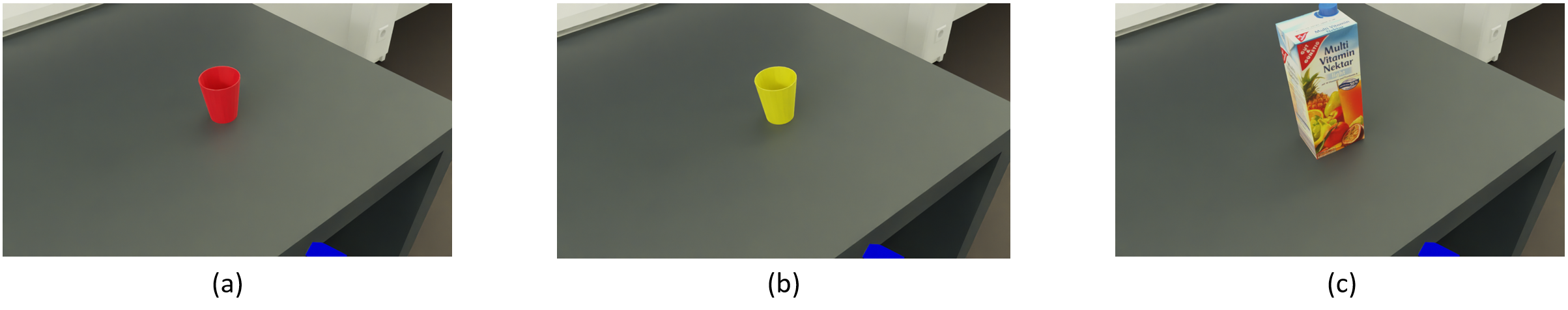}
    \caption{
        Techniques to visually augment experiences using a photo-realistic simulation engine. \textbf{(a)} shows the original scene. \textbf{(b)} shows the same scene but with a different colored object. \textbf{(c)} shows the same scene with a replaced object.
    }
    \label{fig:ltm_augmentation}
\end{figure}

% Blender augmentation
We additionally implemented methods to use the information in the LTM for the generation of new knowledge. Internal simulation is a powerful technique to refine the models used for generalization and prediction in our deep episodic memory. Using a high-quality simulation engine~\cite{blender_2022}, we are able to render past experiences in new scenes or even to simulate the interaction with new objects as shown in~\cref{fig:ltm_augmentation}. This technique is only used to simulate visual data, but we plan to use it for physical interactions as well. The simulated data is finally used along with real data to train the auto-encoders of the deep episodic memory.

In summary, both the WM and the LTM are active parts of the memory system. The WM can adapt its behavior to the current computer load and refines its models to predict information and the LTM actively searches for suitable data representations in order to compress, reproduce, generalize and predict them.

\section{Technical Evaluation}
\label{sec:evaluation}

We evaluated the new memory system empirically. Since all components communicate with and through the WM we begin by evaluating the WM. The WM must support large data transfers while providing a high availability for querying information. Second, we evaluate the LTM. Other than the WM, the LTM must be able to store lots of data which, sooner or later, requires the LTM to compress or even forget information. 
%We measure, how good the LTM converts the information into a compressed representation which also generalizes and allows prediction.

% Let's write single "bytes" out, only abbrev. "GB" and "MB".
All measurements were performed on computers equipped with an 
% We need to allow breaking here, otherwise it will stand out ...
AMD Ryzen~9~5900X with 12 cores, 
$64$GB RAM and a NVIDIA~GeForce~RTX~3060 running Ubuntu~20.04. 
We define three different data objects with an increasing complexity taken from real-use implementations in \armarx.
First, a simple object only containing a single long value (\SI{8}{bytes} of information). 
Second, a moderate object containing a long, a string (with a fixed length of five chars) and a memory link. In sum, this moderate data object has \SI{33}{bytes} of information. 
Finally, a complex object containing memory links, nested objects, an image with resolution $128 \times 128 \times 3$ and several maps with fixed size. 
In total, this object contains  \SI{49.225}{bytes} of information. 
%We will show that the optimizations of the IDF NDArray \dataobject{} greatly influence the conversion and transfer times. 

For each test, we calculated the means and variances over $1000$ samples. Further, we performed each test either with one single data entry or with a batch of $20, 50$ or $100$ entries.

\subsection{Evaluation of the WM}

\newcommand{\microseconds}[0]{\SI{}{\micro\second}}
\newcommand{\range}[2]{$[ \num{#1}, \num{#2} ]$}

We measure the time a producer needs to convert business objects and send IDF \dataobjects{} to the memory and how long a consumer needs to receive and convert the information back after being notified by the memory. %We measure the time needed to convert the auto-generated business objects to IDF and vice-versa. 
Since all components run in different processes and maybe on different machines, the chosen middleware ZeroC Ice~\cite{ZeroCIce} requires time to serialize the information, access the network stack, transfer the information to the remote machine and finally deserialize the information. We treat the middleware as a black box system, thus, we have no influence on the times required for each of those steps. Because our memory is written independent of the chosen middleware, it could also be replaced by \eg ROS, which has lower mean latency values for the transmission of data~\cite{ROS2}.
We will compare the measured results against direct peer-to-peer connections and publish/subscribe channels.

\begin{figure}[tb]
    \centering
    \includegraphics[width=\linewidth,bb=0 300 800 550,clip]{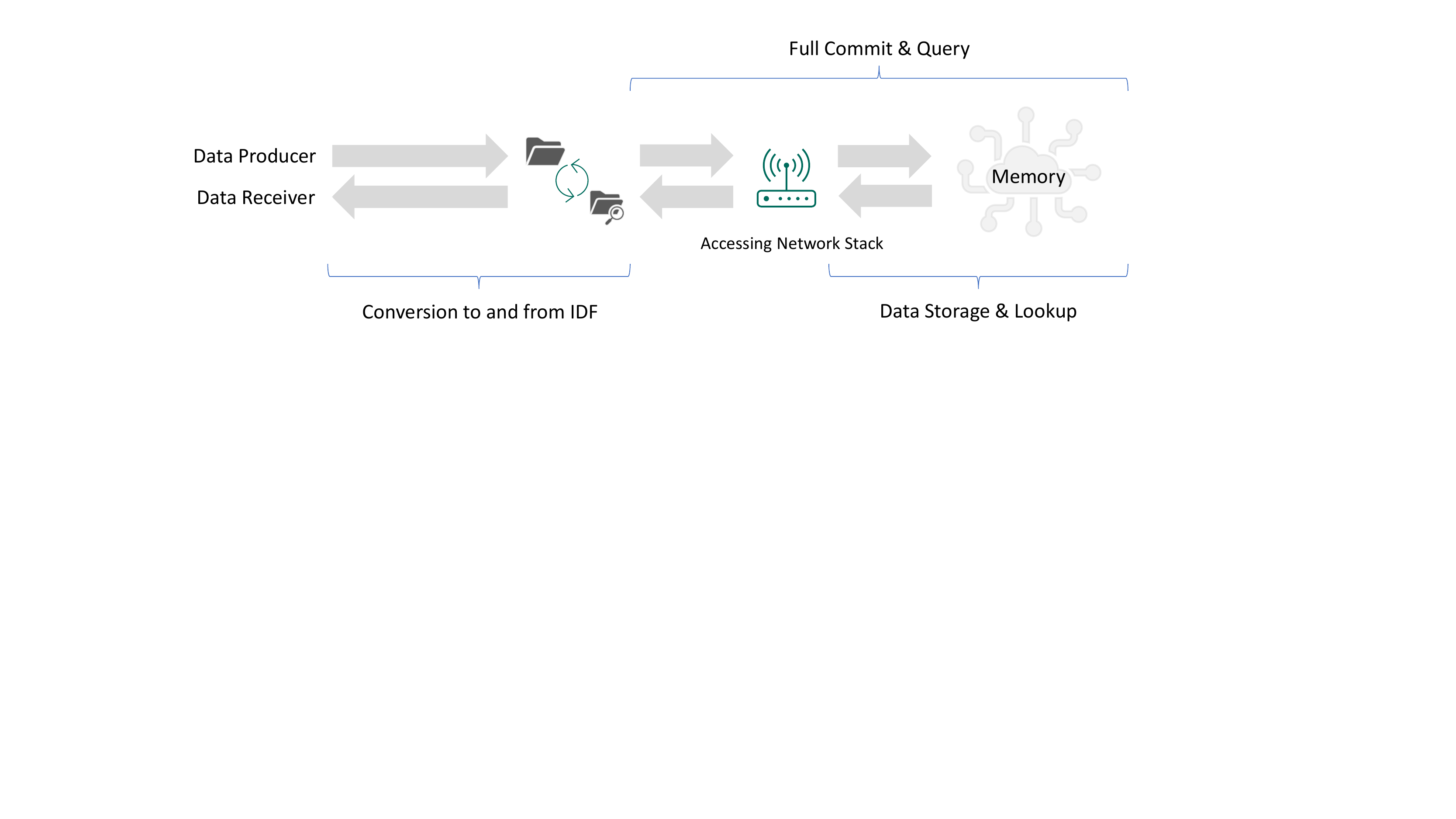}
    \caption{Overview of data transfer to and from the memory from arbitrary components. We evaluated our memory by measuring the times required to commit to and query data from the memory.}
    \label{fig:evaluation:overview}
\end{figure}
\begin{figure}[tb]
    \centering
    \includegraphics[width=\linewidth,bb=0 300 900 550,clip]{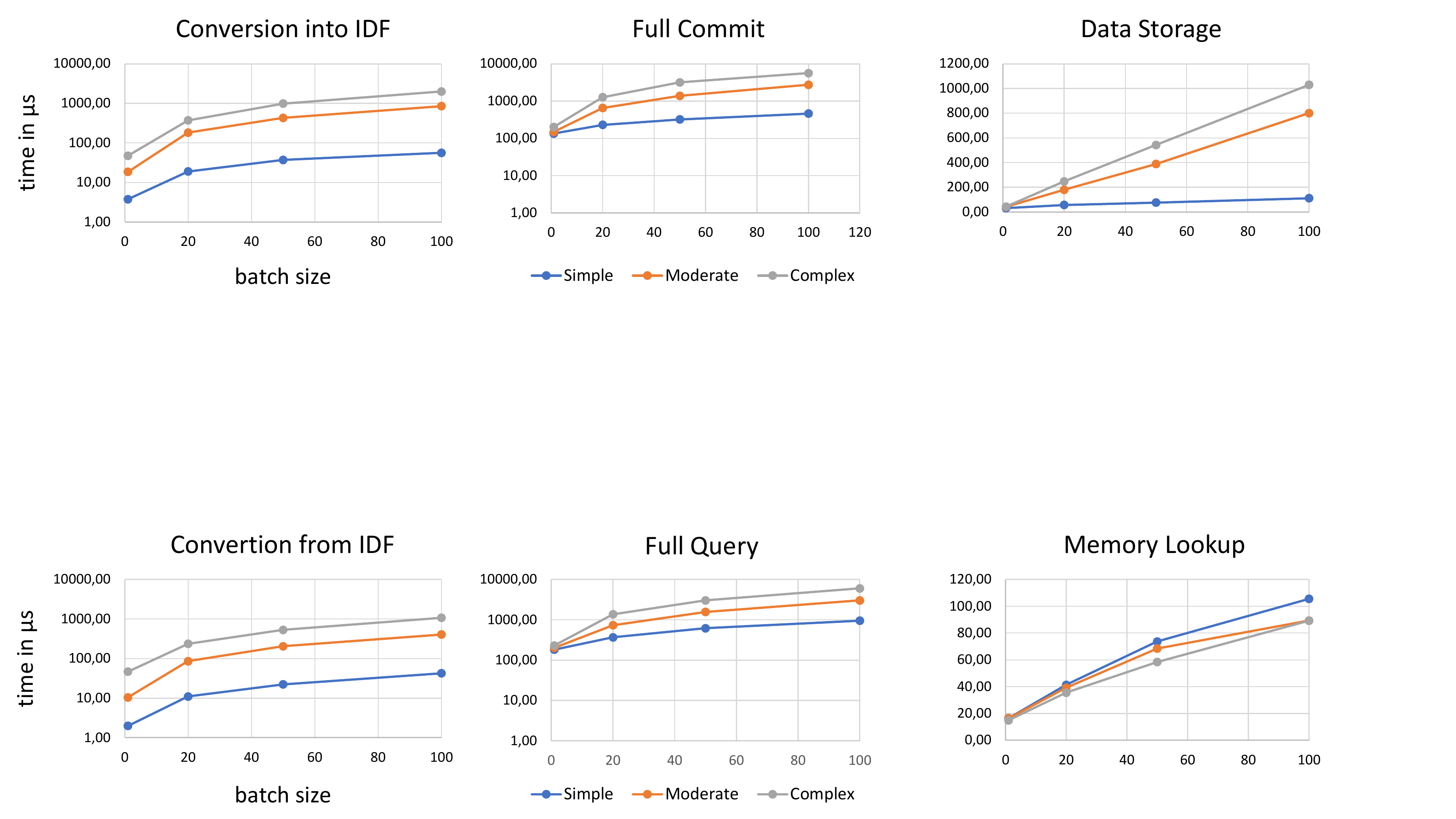}
    \caption{Measured times to send a single or multiple instances to the memory including the time to convert the auto-generated business objects to IDF. We calculated the mean values over $1000$ individual samples per data type and per batch size. Note that the first two charts have a logarithmic scale on the y-axis.}
    \label{fig:evaluation:1}
\end{figure}
\cref{fig:evaluation:overview} provides an overview of how the measured times affect the overall data flow to and from memory. The times to commit information to the memory are shown in \cref{fig:evaluation:1}. 
%To better  In the following we will identify the charts by the objects complexity followed by the batch size, \eg \texttt{SIMPLE\_1} for a simple object with batch size $ = 1$
The diagram ``\emph{Full Commit}'' depicts the times needed to commit IDF \dataobjects{} to the memory, including the network overhead. The diagram ``\emph{Data Storage}'' however only shows the time needed to insert the received information into the memory structure. %If we subtract the mean values from each other, we get an estimate over the network overhead (\eg $\SI{6284.58}{\micro\second} - \SI{1157.64}{\micro\second} = \SI{5126.94}{\micro\second}$).

With increasing complexity the times required to convert or commit large batches of objects increase linearly with their data size. However, when transferring small amounts of data over the network we observe minimum costs for accessing the network stack. Even a transfer of small objects takes at least $100 \microseconds$.
%If we assume a constant network access time for the \texttt{SIMPLE\_1} and the \texttt{MODERATE\_1} experiment, we can calculate the time required to access the network stack $n = \SI{150}{\micro\second}$
We can also observe, that the times to convert and transfer batches of moderate data are closer to the ones of complex data although its data size is closer to the simple type. One has to keep in mind that the biggest part of the complex type comes from an image ($49.152$ bytes) which will be converted into a single IDF NDArray which can be converted and transferred much more efficiently. Apart from the image, the complex type is approximately two times bigger than the moderate type, which can also be observed when comparing the transfer duration of both experiments.

The time to convert a business object to its IDF representation not only depends on the data size -- it also depends on the amount of members %as the converter has to allocate new objects for each member 
and on the batch size. %as the converter has to reallocate the writer object for every instance
This  is why the conversion of a moderate type is approximately $15$ times slower than the simple one. The moderate and complex objects almost have a similar number of members. % Further, from the experiments we can see that it is easier for our converter to convert a batch of information than converting
%
%That commit times increase linearly with data size becomes particularly clear when comparing commit times of experiment \texttt{SIMPLE\_100} and \texttt{MODERATE\_100}. The time to commit information of experiment \texttt{MODERATE\_100} to the memory via the network is approximately $4.88$ times higher than in experiment \texttt{SIMPLE\_100}. This is expected behavior as the size of the moderate type is $4.125$ times bigger than the simple one. 
%The time to commit complex data to the memory is more influenced by big members than the IDF conversion. Thus, \texttt{COMPLEX\_100} requires around three times longer than \texttt{MODERATE\_100}. Nevertheless, when keeping in mind that a complex object contains $1490$ times more bytes than the moderate one, we observe that the utilization of byte-arrays in IDF for large members greatly affects the transfer and conversion time. 
%
The times to arrange information in the memory only depend on the data size. In our experiments, we used a single memory server with already $1000$ entries to ensure that all insertion times are comparable. Further, a memory with already existing entries is closer to real-use scenarios. While we can observe a small static overhead to access the memory structure and to insert commits of experiments, the measured times for the batched experiments prove the aforementioned dependency on the data size.

\begin{figure}[bt]
    \centering
    \includegraphics[width=\linewidth,bb=0 0 900 200]{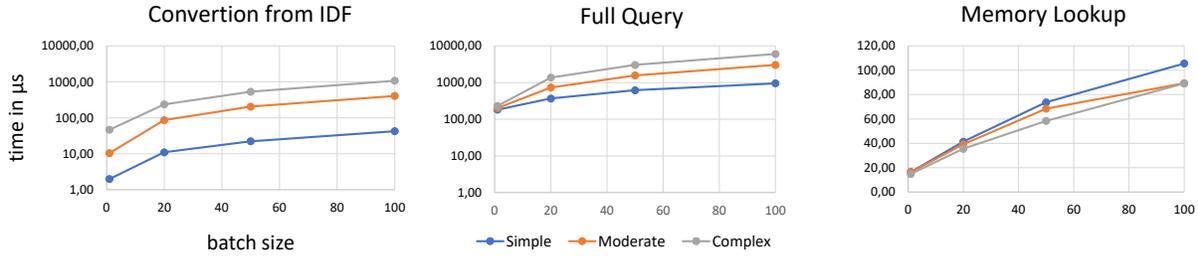}
    \caption{Measured times to query a single or multiple instances from the memory including the time to convert the transferred data in IDF to the auto-generated classes. Again, we calculated the mean values over $1000$ individual samples per data type and per batch size. Similar to \ref{fig:evaluation:1}, the first and the second chart have a logarithmic scale on the y-axis.}
    \label{fig:evaluation:2}
\end{figure}

Once the memory receives new data, it will broadcast a notification to all subscribers. The subscribers may then query the information to receive the new data. The times required to query information are shown in \cref{fig:evaluation:2}. The chart ``\emph{Full Query}'' shows the times it takes to query and receive data from a memory server without converting the data from IDF to its business object representation and
%Those time include the network overhead. 
``\emph{Memory Lookup}'' shows the time needed to search for a specific entry in the memory. 
%The difference of both values again provides an approximate to the network overhead of our middleware.

Similar to committing information via the network we observe that querying information depends on the data size and the batch size while there is a static overhead for accessing the network stack. 
%Again, the overhead comes clear when comparing the measurements with batch size $1$. If there were no overhead, we would expect a similar increase as for the measurements with batch size $100$ (approximately~$\times 4$ from simple to moderate and~$\times 2$ from moderate to complex).
We find out that the time required to search for instances in the memory is almost similar over all experiments with the same batch size because the memory only returns references to the stored information. %Since the memory has to return more references for the batched experiments, the query times are slightly longer.
During a query, the data is automatically converted back to its business object representation. Similar to the opposite conversion we find a similar dependency on the number of members and the data size. 
%The reason for this is that no new objects are created when reading from the IDF representation. 
% The increased time to convert the complex data in experiment \texttt{COMPLEX\_1} can be explained by the fact that we copy the image information twice (thus, having a greater influence on the full conversion time). This redundancy gets compensated in experiment \texttt{COMPLEX\_100} through the accelerated transfer time of the IDF NDArray. 

\begin{figure}[bt]
    \centering
    \includegraphics[width=\linewidth,bb=0 250 800 550,clip]{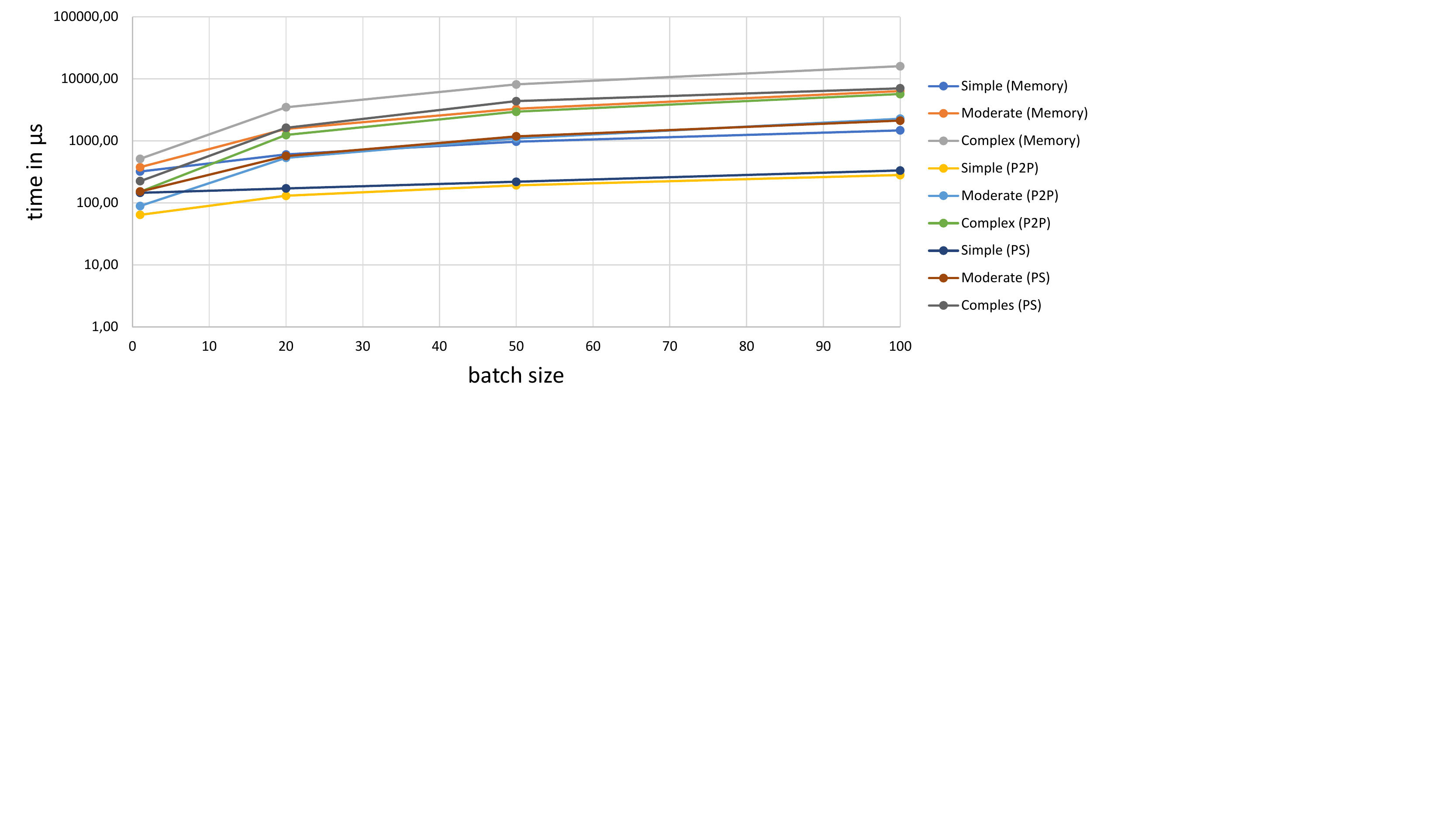}
    \caption{Averaged age of data as difference of timestamps between data reception and production,  either using the memory, peer-to-peer (P2P) connections or publish/subscribe (PS) channels provided by the  middleware ZeroC Ice.}
    \label{fig:evaluation:3}
\end{figure}

To conclude the results of our first experiment, we compared the full times from producing information to receiving that information in the consumer component with direct peer-to-peer (P2P) connections and publish/subscribe (PS) channels managed through a centralized service as shown in \cref{fig:evaluation:3}. P2P times can be seen as a theoretical lower bound for sending data in a distributed system. A PS service is more comparable to the memory as it receives data from producer components and forwards it to an arbitrary number of subscribers. However, it lacks the ability to access a history of data or to actively request data.

% \begin{table}[ht]
%     \centering
%     %\resizebox{0.5\textwidth}{!}{
%     \begin{tabular}{ccccc}
%         \toprule
%         \textbf{Type}
%         & \makecell{\textbf{Batch} \\ \textbf{Size}} 
%         & \makecell{\textbf{Total Time} \\ \textbf{(memory)}} 
%         & \makecell{\textbf{Total Time} \\ \textbf{(p2p)}} 
%         & \makecell{\textbf{Total Time} \\ \textbf{(ps)}} 
%         \\
%         \midrule
%         Simple & 1 & \num{364.67} & \num{46.68} & \num{177.17} \\
%         %\hline
%         Moderate & 1 & \num{417.54} & \num{87.89} & \num{190.69} \\
%         %\hline
%         Complex & 1 & \num{548.27} & \num{137.98} & \num{249.58} \\
%         %\hline
%         %\hline
%         \midrule
%         Simple & 100 & \num{1494.83} & \num{288.319} & \num{310.876} \\
%         %\hline
%         Moderate & 100 & \num{6393.54} & \num{1242.47} & \num{2029.96} \\
%         %\hline
%         Complex & 100 & \num{15238.3} & \num{5978.74} & \num{7004.56} \\
%         \bottomrule
%     \end{tabular}%}
%     \caption{Measured times (in \microseconds, averaged over $1000$ tests) to send data over the network from a producer to a consumer. We compare the times needed to send data via the memory, via direct peer-to-peer (p2p) connections and via a publish/subscribe (ps) service, provided by our middleware ZeroC Ice.}
%     \label{tab:memory_evaluation:3}
% \end{table}

The memory performs worse of the three, but it offers important features for a cognitive system, such as being able to query information from the past or to provide general data-driven services to clients. Further, this evaluation should be seen as a worst-case scenario. Usually, producer, memory and consumer processes are not implemented as independent processes or running on different machines. Producers often implement a memory server, so that data does not have to be transferred over the network. The times measured here can therefore be halved in many cases and by a well-chosen distribution of components.
Nonetheless, even latency times measured here for moderate data types allow data to be retrieved from the memory at a maximum frequency of over \SI{160}{\hertz}, which is enough for state-of-the-art model predictive control algorithms such as in \cite{STORM}. For the few components in our software framework that require accessing the data more frequently, such as self-collision checks and the emergency stop, side channels (P2P and PS channels) are still possible and valid.

\subsection{Evaluation of the Deep Episodic Memory}
In the long-term memory, it is important to find models with a good compression rate, generalization ability and prediction ability which we will briefly evaluate in this section. For this evaluation, we recorded five experiments on the real humanoid robot \armarIII. Similar to~\cite{Baermann_2021}, the recordings include stereo camera RGB images (with a resolution of $640 \times 480$ pixels at a frequency of \SI{20}{\hertz}), the robots pose and configuration including the positions and velocities of \SI{43}{\dof}, action information, plans, object instances and their locations. The robot was instructed to perform simple pick-and-place tasks in a kitchen environment.

If unfiltered, the LTM receives $35.665$MB of data per second. Images are the largest part of the data. The online compression used for the recordings reduced the amount of required disk space to $0.213$MB per second (reduction by 99.4\%). First, the input data was downsampled to a maximum frequency of \SI{5}{\hertz}. Further, images were compressed using PNG compression and all data was converted into a binary format using ZIP compression. Offline compression using Auto Encoders further reduced the required disk size to
$0.00383$MB per second (which represents an additional reduction by 98.2\%).

\begin{figure}[bt]
    \centering
    \includegraphics[width=\linewidth,bb=0 100 950 500]{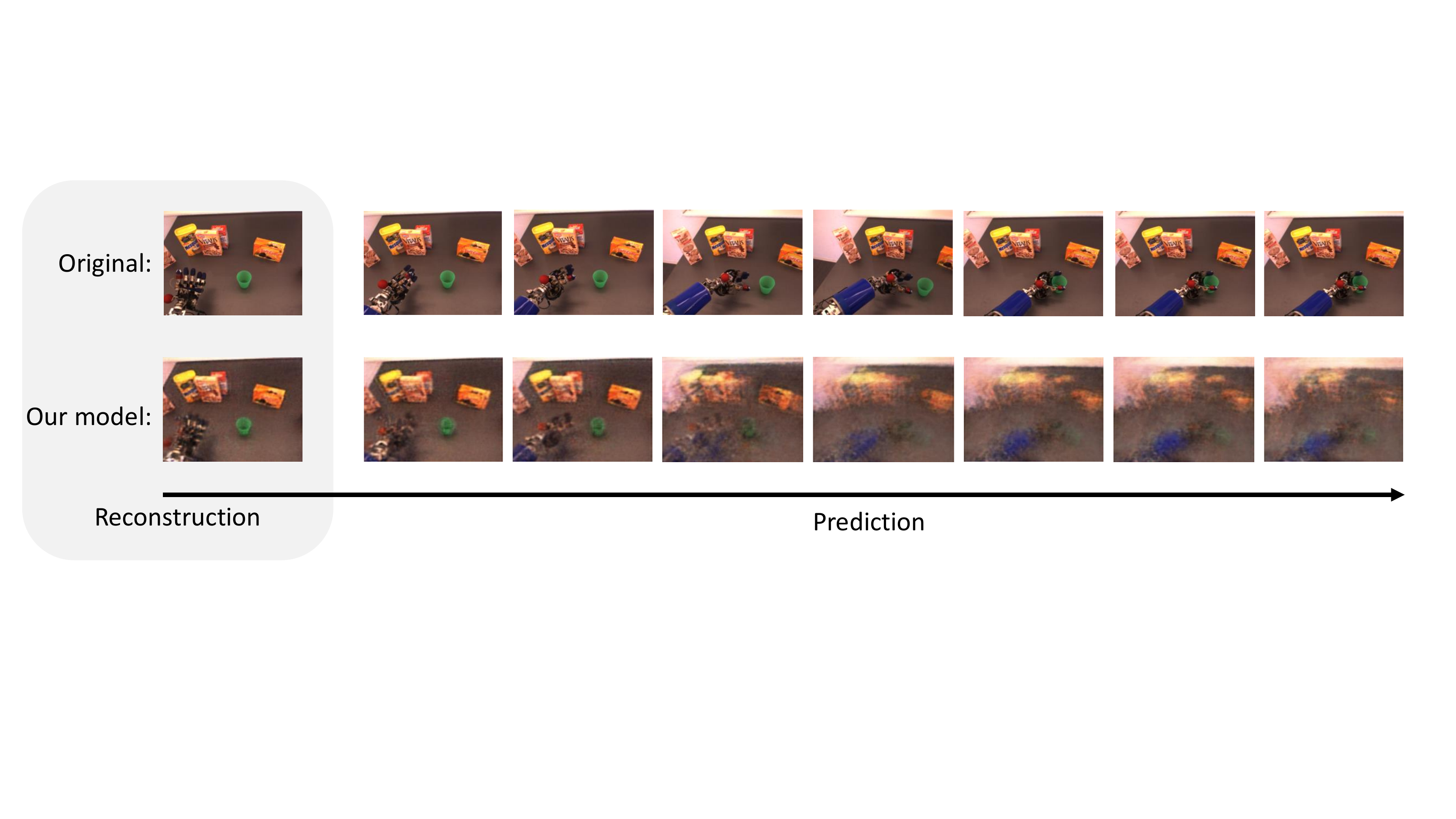}
    \caption{Reconstruction and predictions of the next few frames returned by our deep episodic memory. The first original image is encoded into a latent space. A decoder model reconstructs the image or predicts the next frames.}
    \label{fig:ltm_evaluation_example}
\end{figure}

For the quantitative evaluation of the network, a leave-one-out cross-validation was performed to obtain $5$ folds of training and testing sets. To compare the results to our previous work~\cite{rothfuss2018}, we will concentrate on visual information and calculate the PSNR~\cite{PSNR} as shown in \cref{tab:loocv_results} for the prediction and for the reconstruction. Because our deep episodic memory automatically instantiates one AE per entity, other modalities can be reconstructed and predicted in the same way. 
%During our experiments, however, images were the most complicated data due to their size and dimensionality. 
The results show, that our multimodal model outperforms the previous work in terms of reconstruction and prediction although the original work was pre-trained using large datasets. Still, the performance to predict frames decreases with increasing horizon. This is natural, as the uncertainty increases. Examples of reconstructed and predicted visual experiences are shown in \ref{fig:ltm_evaluation_example}. 
%Those examples clearly show, that our network benefits from the ability of the WAE to avoid overlapping distributions during training. During the first predicted frames, the VAE model has difficulties to decode the chosen distribution, resulting in almost the same image for three prediction steps. Further, images reconstructed or predicted from the WAE model are less noisy, due to the missing sampling step from the latent space.

With this work, we want to briefly introduce the intermediate results of our research regarding representation learning in the robot's long-term memory. We are still working on a final and all-encompassing evaluation of the presented system with many different modalities and real application examples. 

\begin{table}[ht]
    \centering
    %\resizebox{\textwidth}{!}{
    \begin{tabular}{cccc}
        \toprule
        \textbf{Iteration}
        & \makecell{\textbf{Reconstruction}}
        & \makecell{\textbf{Prediction}\\ \textbf{of next frame}}
        & \makecell{\textbf{Prediction}\\ \textbf{of over next frame}}
        \\
        \midrule
        1 & \num{46,8285912} & \num{33,67589698} & \num{28,30841852} \\
        %\hline
        2 & \num{48,63735492} & \num{32,26709011} & \num{34,61204175} \\
        %\hline
        3 & \num{44,22040739} & \num{32,05658062} & \num{30,04516065} \\
        %\hline
        %\hline
        4 & \num{52,72365549} & \num{35,52542389} & \num{30,98289146} \\
        %\hline
        5 & \num{48,21215029} & \num{33,11685694} & \num{28,99858607} \\
        %\hline
        \bottomrule
    \end{tabular}%}
    \caption{PSNR values (averaged) of the test set during a leave-one-out-cross-validation of the generalization network used to compress visual information in the long-term memory. The generalized latent representation offers the possibility of reproduction, production and finding of similar instances as described in~\cite{rothfuss2018}.}
    \label{tab:loocv_results}
\end{table}

\section{Case studies}
\label{sec:casestudies}
% alternative titel:

%\todo{Hier viel mehr. Ggf könnte man das sogar als eigenes Kapitel einbauen, da es über die Implementierung hinaus geht. Dafür sollte es  aber minimum 2 Seiten füllen}

%\todo{ Ich denke wir sollten das hier mehr wie eine Evaluation aufziehen. 1. genauer zeigen wie das Memory als Mediator zw. High und Low-Level dient. Wo die ... }

In this section, we will showcase different use cases of varying complexity, showing how different components are supported by the memory system and how processing pipelines can structure their data flow to make the best use of it.
The following examples should be seen as different views on one and the same robot's cognitive architecture, \eg there is only one \memid{Robot State} memory managing information related to the robot pose and configuration which can be used by any other component.

In our architecture, executable robot skills and behaviors are managed through a \memid{Skill} memory. It connects executable codelets with memory. In the following, we  describe this memory-driven skill framework using the example of how a high-level speech command is being stored and processed within the memory architecture resulting in the desired robot's low-level behavior. Here, we  focus on two different aspects (1) high-level symbolic skill representation and (2) sub-symbolic skill execution. The former will explain how arbitrary skills can be registered to the memory. The latter will explain how the memory system listens to certain conditions to instantiate and execute skills.

% how does memory system work from a user's perspective?

% aktiv
% assoziativ
% introspektiv
% distributed
% data access
% multi modal
%While the first use-case only focused on querying knowledge from the (long-term) memory and externalize it, another important aspect of robotics is to gather knowledge from interaction with the environment.

\subsection{Skill Representation}

\begin{figure}[ht]
    \centering
    \includegraphics[width=0.7\textwidth,bb=50 140 900 600,clip]{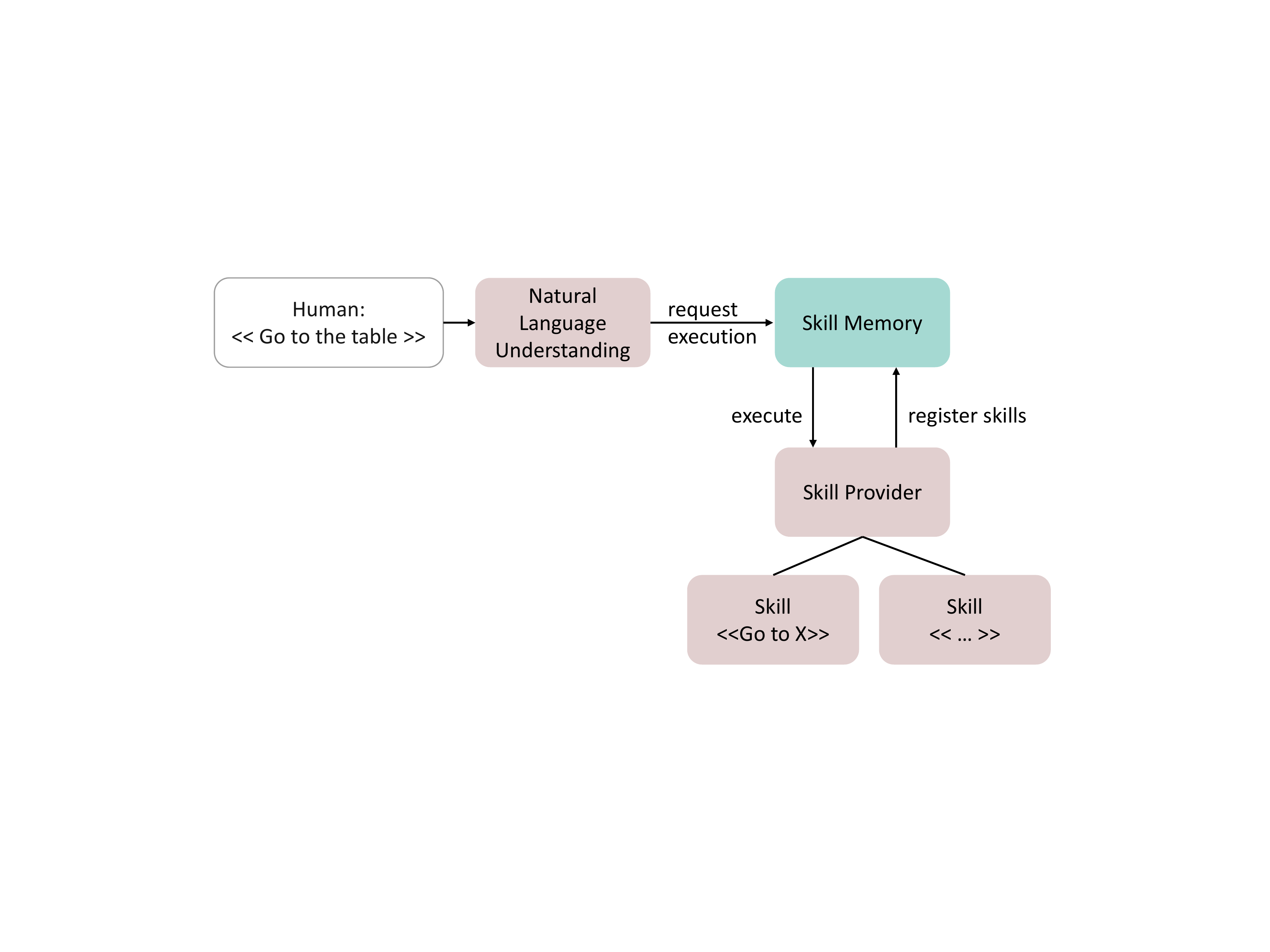}
    \caption{The connection of executable skills to the memory system. Skills can be triggered by \eg speech.}
    \label{fig:use_cases:navigation:skill_framework}
\end{figure}

\emph{Skill Providers} register information related to executable codelets, so called \emph{skills}, to the skill memory as shown in Figure~\ref{fig:use_cases:navigation:skill_framework}. Skills that are available during start-up are taken from the prior knowledge (PK) of the robots' memory. However, skills can be added, removed, enabled or disabled at any time. An instance of a skill in the memory incorporates a symbolic description that itself links to executable code. There can be a variety of skill providers which register skills in this memory. Skill providers thus group corresponding and related skills. Depending on the definition of the skill, it can be parameterized both using symbolic (\eg{} ''Go to the table'') and sub-symbolic data (\eg{} ''Go to position (10,10)''). A skill may be linked to conditions that trigger its execution. The simplest condition may be that the execution of a skill is directly requested. 
However, one can also map complex conditions, such as in response to a certain events such as a recognized object in the case of grasping.
Once all conditions of a skill are fulfilled, the memory covers the instantiation and execution of the executable codelet of the skill. An event that the skill is executed is stored in the skill memory. The result of the execution (\eg whether it was successful, stopped or unsuccessful) is stored in the memory as well. Intermediate results are usually associated with the execution of the skill. In the following, we will further describe how a skill internally uses the memory for its operation using the examples of navigation, object localization and knowledge verbalization. 

\subsection{Navigation}

% -> aktiv
% assoziativ
% -> introspektiv
% distributed => will be described in UC "object detection and localization" 
% -> data access
% -> multi modal

\begin{figure}[ht]
    \centering
    \includegraphics[width=0.9\textwidth,bb=50 50 800 600,clip]{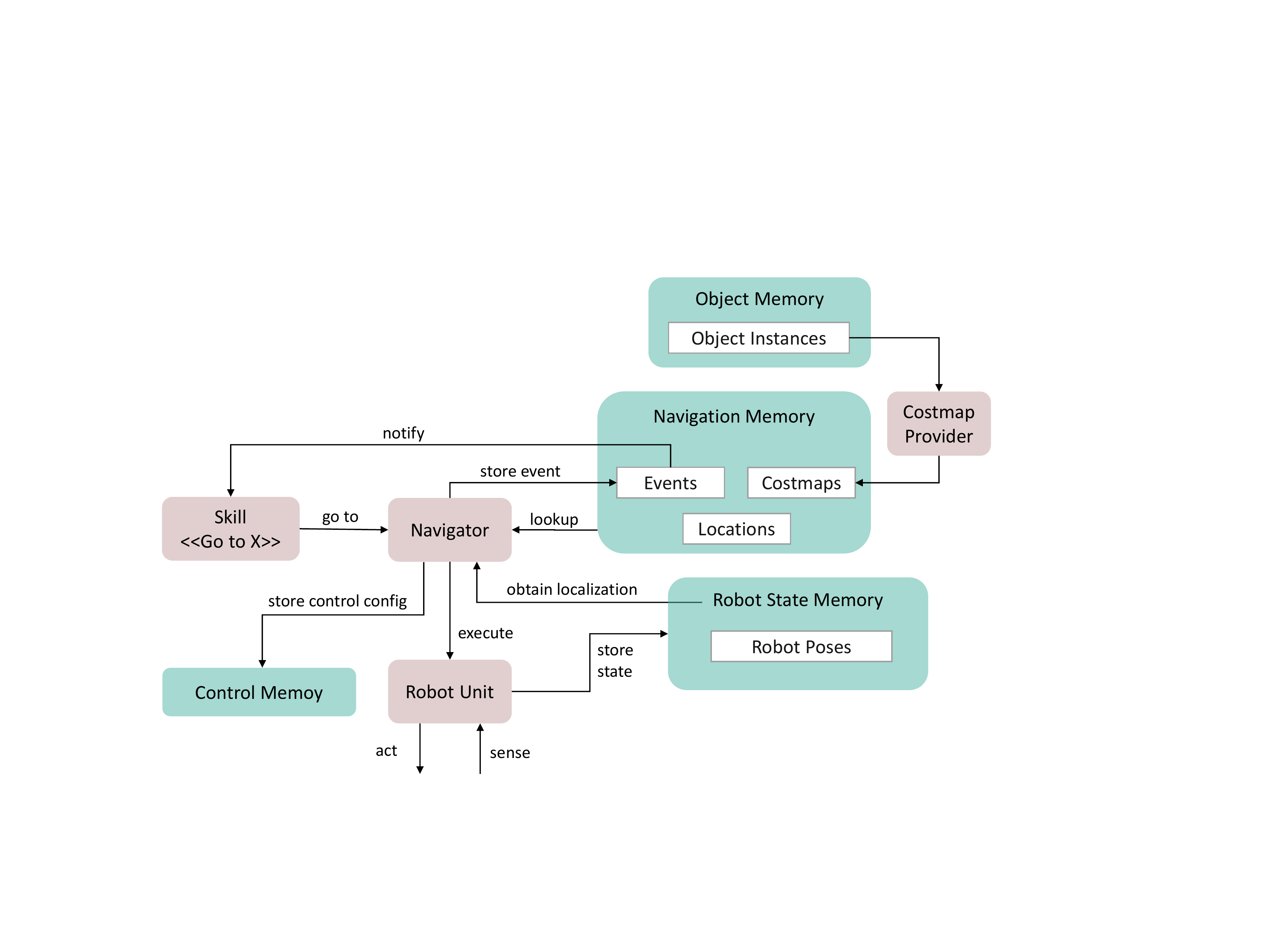}
    \caption{Initial planning phase during the navigation skill execution.}
    \label{fig:use_cases:navigation:initial_planning}
\end{figure}

%While the first use-case only focused on querying knowledge from the (long-term) memory and to externalize it, another important aspect of robotics is to change the state of the robot and to gather knowledge from interaction with the environment.
Robot navigation is realized through navigation skills.
Once the user requests the robot to ''Go to the table'' such a navigation skill gets executed. It will access the navigator service as depicted in~\cref{fig:use_cases:navigation:initial_planning}. This service resolves the Cartesian position of the location, plans a collision-free path, executes the movement and reacts to dynamic changes in the environment. 
Once the planning phase ends, the result will be stored in a specific \memid{Navigation} memory.
%Through the distributed memory system, all of the required information can be retrieved. 
During planning, accessed data can be multi-modal such as named locations and cost maps 
%(in the navigation memory}) 
needed for path planning, as well as the robot's global pose. In addition, components such as the cost map provider ensure that the navigation memory actively updates itself, as a change to the object memory can trigger an update to the distance cost map.
 The planning result is associated with the corresponding skill execution request. This association may be used for question answering (\eg ''Why did you move to the table'') or for debugging purposes. Afterwards, the navigator service continues with the execution phase. The skill waits for the robot to reach the desired location but can also react to unforeseen events such as the robot getting trapped due to surrounding humans. 
 
As for any memory server, clients subscribe to receive updates on memory segments. Thus, other components can also listen to the \memid{Navigation/Events} core segment and react to events such as the robot starting to move or there is a planning failure.

%In this case, the skill is notified of events in the \memid{Navigation/Events} core segment. 
%The client is then notified of the current status of the navigation request, e.g., that the robot starts to move, that there is a planning failure, or that the robot has reached the goal. 
%When low latency is critical, a command, \eg{}, following a desired trajectory, can be sent directly to another component over a peer-to-peer connection. In particular, this is the case for control commands. As this command can be described using our IDF, the data can easily be also be stored in the memory after it was sent to the component for further introspection.

\begin{figure}[ht]
    \centering
    \includegraphics[width=0.9\textwidth,bb=0 0 1100 600, clip]{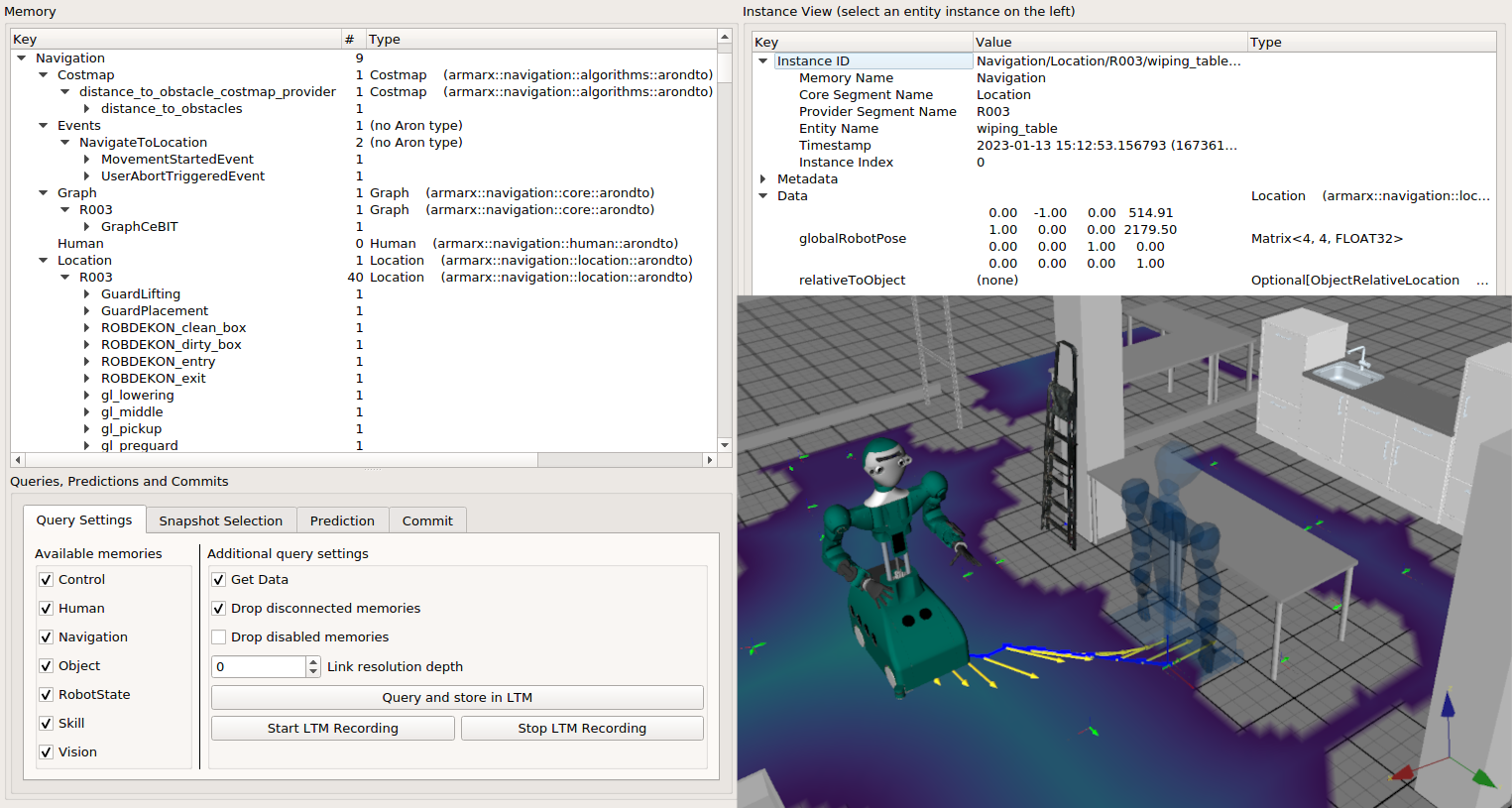}
    \caption{The memory GUI and the 3D visualization.}
    \label{fig:use_cases:navigation:memory-gui}
\end{figure}

All information in the memory system is fully introspective for the system but also for the user. As an example, the state of the memory system after the initial planning phase is shown in Figure~\ref{fig:use_cases:navigation:memory-gui}. On the left, the GUI shows the content of the \textit{navigation memory} as a tree view. Each entity instance can be investigated as shown on the right. In this example, the view shows the robot's platform location associated with the table. On the bottom right, the content of the memory is visualized in 3D. 

\subsection{Object Detection and Localization}

% % Object pose requesting, etc.

% multi-modal
% active
% distributed
% associated

\begin{figure}[ht]
    \centering
    \includegraphics[width=0.9\textwidth,bb=0 150 1100 650, clip]{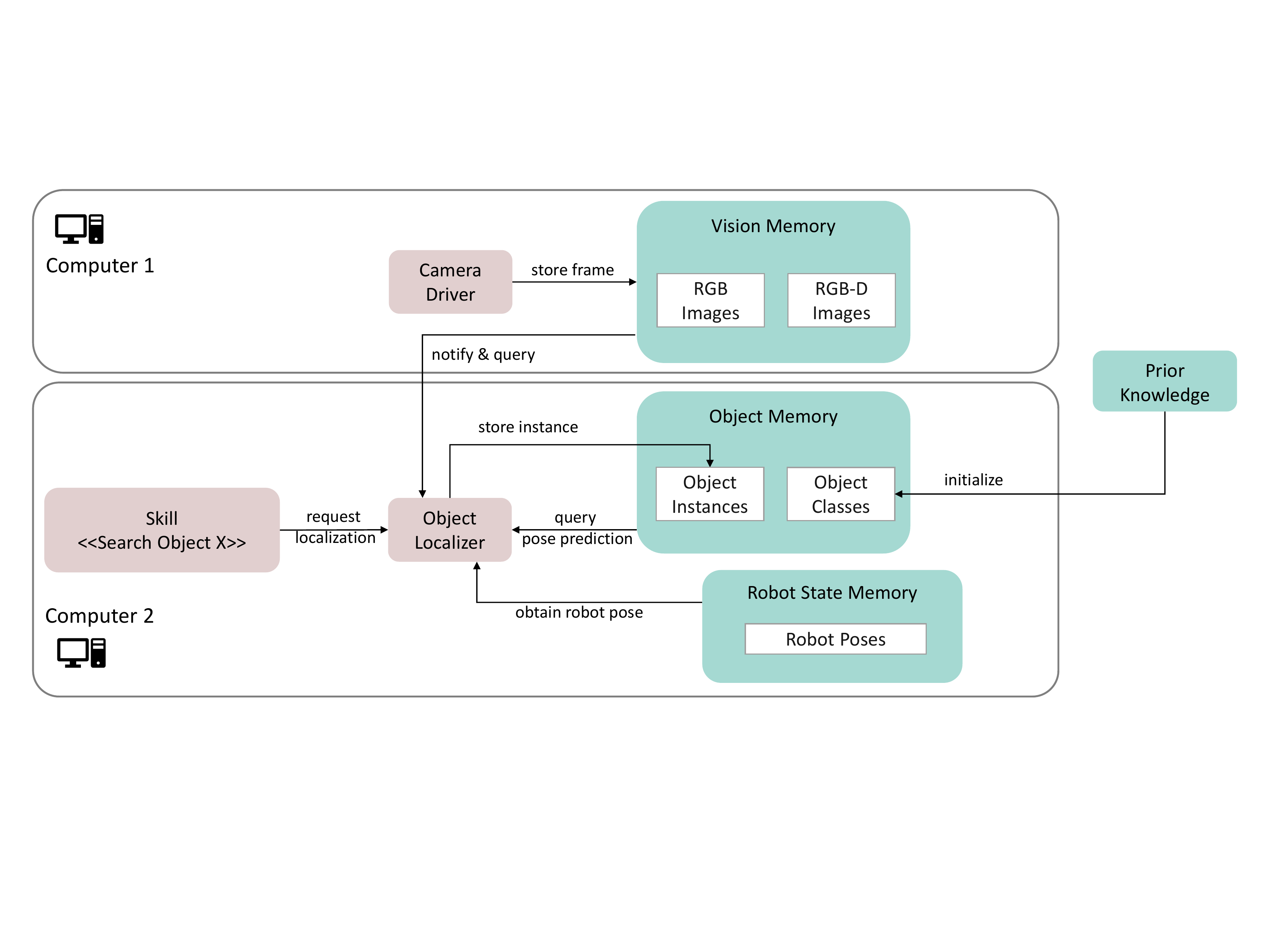}
    \caption{Integrating an object localization component into the distributed memory system.}
    \label{fig:use-cases-object-localization}
\end{figure}

Another important ability of a robot is to detect and localize known or unknown objects in its environment.
In this example, we consider the task of 6D localization of known, textured objects based on RGB or RGB-D images.
% as in, \eg, \citet{Azad2009CombiningHarrisInterest,Pauwels2015SimTrack, He2021FFB6D}.
% %
Consider the processing pipeline in \cref{fig:use-cases-object-localization}.
A camera driver component reads data from a hardware camera.
In each frame, it commits the RGB and depth images to two separate segments in the \memid{Vision} memory.
% %
If enabled by the user (\eg by running an object localization skill) an object localization component listens to updates of the \memid{Vision/RGB} and \memid{Vision/RGB-D} segments to get notified about new images. 
When notified, it queries the latest images from both segments.
In addition, it uses multi-modal related prior object information such as 3D models, likely positions or pre-extracted visual features from the \memid{Object/Class} segment which itself is initialized from the prior knowledge.
The localization component then applies its internal method to detect objects and estimate their poses. 
Finally, it commits the new observations to the \memid{Object/Instance} segment, which holds the current state of objects in the scene. Object poses are associated with the images they are calculated from. Additionally, they are linked to the current robot pose. The latter allows for analyzing how the object was placed relative to the robot and how the robot interacted with the object retrospectively. For efficiency, camera drivers and the vision memory are running on a separate machine, equipped with special-purpose hardware for computer vision. Prior knowledge is equally available on all machines. 

Some 6D localization components benefit from an initial guess \eg regarding the pose of a moving object. Our memory supports those components by providing methods to predict information. This functionality is not limited to object poses -- if a memory server only contains data of similar type it can predict the numeric elements. Prediction is realized on two levels: (i) fast methods only using information stored in the WM with updates each time new data enters the WM, and (ii) more complex systems using the information stored in the LTM as described in \ref{sec:implementation:system:ltm}. Requesting a prediction from the memory is done by querying a future timestamp. Depending on the memory server's configuration and the amount of received data, the memory decides to either use the fast methods of (i) or more complex systems of (ii)
%\todo{das ist nicht klar xxxxxxxxxx} 
to generate a prediction. Refining the models based on an evaluation of how good this estimation has been is not part of this work.

\subsection{Verbalization of Episodic Knowledge}
\label{sec:casestudies:verbalization}

% Wo introspektiv, wo aktiv, ...
% Wie latent spaces, warum sinnvoll
% Übergang High und low-level
% Multi-modal integrieren - check
% aktiv (inkrementell), introspektiv
% episodic - check
% long-term - check
%(explainability)

Verbalization of episodic knowledge is fundamental for natural human-robot interaction. This ability requires the robot to understand the user's query, access its memory and finally generate a response. In our previous work~\cite{Baermann_2021} we presented a system that is able to answer user questions about the past, such as ''Where have you seen the red cup?'' or ''What did you yesterday?'' by using the knowledge stored in the robots deep episodic memory. Further, the robot is able to verbalize important events such as successes or failures of actions, \eg ''I tried to put the milk in the fridge but I failed to open the door''.

Technically, a special verbalization skill listens to the \memid{Speech} memory server. This memory server contains recognized spoken text as simple strings. Every time a spoken text is recognized and this update is broadcasted by the memory, the verbalization skill queries the latest recognized text and interprets this text as a verbalization command. Thus, we assume that the skill has been activated so that the robot is in verbalization mode. 

The user query is passed into a neural network which internally applies attention~\cite{Xue_2020} on the latent instances stored in the robot's deep episodic memory. Plain text knowledge of the WM is used for verbalization. The latent representations are learned unsupervised and beforehand as described in section \ref{sec:implementation:system:ltm}. As the robot should be able to verbalize symbolic and sub-symbolic knowledge our verbalization component requires access to multi-modal data sources. In our experiments, we granted access to executed plans and skills 
%as part of the \emph{skill memory}
, object information 
%in the \emph{object memory}
, robot configuration and positions 
%in the \emph{robot state memory} 
and visual information. %in the \emph{vision memory}. 
A decoder network finally converts the latent representation of the user query and the latent representations of the attended episodes into a response from the robot. The generated response is again stored in a separate segment of the speech memory of the robot.
The deep episodic memory depends on its content, as its weights may be updated with each new instance which gets consolidated from WM to LTM. Thus, this part of our memory cannot be seen as a simple passive storage device, fulfilling the requirement of an active memory. Further, learning from data instances requires an understanding of the data. This would not be possible without the aforementioned introspectability of our data representation.

% https://drive.google.com/file/d/1zL8NSCJJTrTw897tkxc5hxc0w_6umzLn/view?usp=sharing

\section{Conclusion and Future Work}
\label{sec:conclusion}

In this work, we identified several characteristics of a memory system for a robot cognitive control architecture and revealed details about the functionality and representation of structures in the memory. We also described the implementation of the memory system in our robot software framework \armarx. 
%Technical and conceptual weaknesses of our previous implementation of the memory \cite{Vahrenkamp_2015, Waechter2018}, have let us identify those several characteristics that are essential for a memory for complex humanoid robot systems
%Some of the requirement are identical to what is proposed in related works but others are new. 
Our new memory system fulfills these requirements in the following way:
\begin{itemize}
    \item It is an \emph{active} memory. The working memory (WM) adapts its behavior based on the current computational load and learns from incoming data and the long-term memory (LTM) adapts its ability to predict and generalize to the given data.
    \item It is \emph{multi-modal}. The memory has no constraints against the input data. Everything can be stored and encoded, no matter if it is symbolic or sub-symbolic information.
    \item It is inherently \emph{episodic}. All information is stored episodically all over the memory -- even semantic information.
    \item It supports an \emph{associative} structure. Entities can be linked to other entities of the same or even of different modalities supporting reasoning and explainability. 
    We believe, that the knowledge of how information is connected is as important as the information itself.
    \item The proposed data representation is \emph{introspective}. This allows the memory to investigate the data and to use it for learning and development, to check constraints and to ensure consistency or even to change information for simulation or augmentation.
    \item Due to the fact that our robots have several special-purpose computers, our memory follows a \emph{distributed} design approach. All subsystems of the memory (memory servers) may run on different machines leading to reduced network traffic and increased response times. A special centralized component, the \emph{Memory Name Service}, manages the connections to all memory servers.
    \item Programmers can choose to either listen to memory updates or to actively poll the memory. Both access types have benefits and drawbacks which depend on the production frequency of data sources. Enabling the memory to support both access types makes the memory more \emph{access-efficient}.
    \item Finally, as robot systems are not online all the time and they have limited storage, the memory must be able to efficiently store information in the \emph{long-term}. 
\end{itemize}

We showed how the data is structured in distributed memory servers, how the data can be accessed and  passed to client components. Each memory server contains a long-term memory back end, which allows the memory server to store knowledge persistently. Whether and when data is moved to the LTM is determined by internal statistics, or by the user. The LTM has the task of storing the data as efficiently as possible. For this purpose, it makes use of various compression methods. In an offline compression step, data is converted into a generalized format that allows the prediction and comparison (in terms of distance measuring) of snapshots using machine learning. An introspective yet efficient data representation allows inspecting data instances and eases communication with the memory through code generation. 

The performed evaluation shows that this memory implementation increases transfer times of data from producer to consumer components in comparison to peer-to-peer transfer due to the new features. Nonetheless, due to optimizations in knowledge representation of large data types, even those increased times are acceptable for robotic applications.

Further, we presented three use cases explaining how the implementation of these requirements influences the way we develop  and evaluate our software. Nonetheless, the system is by far not complete as the development of a novel cognitive architecture (from which the memory is the key element) usually requires years. So far the memory accepts only a temporal structure and temporal requests, however, human episodic memory manages knowledge spatially and contextually~\cite{Dere_2004}. Allowing spatio-temporal and contextual access to information will allow more efficient retrieval of data. 
Regarding the efficient management of data, our LTM currently only provides rudimentary tools to filter incoming data and remove existing instances. As a design choice, the WM data is not filtered or encoded at all. During our experiments, we did not reach the limits of storage capacity but we believe that we need  a wider range of filter and removal algorithms to manage the data in the LTM more efficiently, especially when considering lifelong learning and 24/7 operating modes. Thus, we are also working on extending the internal statistics (e.g., using saliency models) to provide a larger groundwork for filtering and forgetting algorithms. 
Moreover, it is currently not possible to incrementally refine the models found for learning representations in the LTM. We either generate a new model only for the experiences of the current work cycle or we remove the models we have, concatenate the decoded experiences and train a new model with this dataset extended by the current work cycle. The latter approach is less space-consuming but also carries the decoding error into the next model. Incremental approaches are desirable. 

Last but not least, there is a lack of methods to evaluate and learn from the stored data. In particular, it would be interesting to compare the memories of several robots and make them available to each other (comparable to what is done in \cite{waibel2011}). Shared memories, with other robots or with humans, also open up questions about security and privacy which should already be addressed at memory level. For this reason, we recently started working towards privacy-aware memory systems and cognitive architecture  \cite{Bayreuther2022}.

The planned work mentioned above represents only a small subset of the further development of the memory system of our cognitive architecture.
%Nevertheless, every single point is a complex research topic on its own. 
We believe that the architecture we have chosen, based on the identified requirements of a memory system, is general and at the same time specific enough to support the development of cognition-enabled and AI-driven robotic applications on complex robot systems such as humanoid robots.

\section{Acknowledgements}

%\ackReallabor
\ackReallaborTERRINet

%\ackTERRINet

\clearpage
\bibliographystyle{unsrt}  
% FRONTIERS
%\bibliographystyle{frontiersinSCNS_ENG_HUMS} % for Science, Engineering and Humanities and Social Sciences articles, for Humanities and Social Sciences articles please include page numbers in the in-text citations
%\bibliographystyle{frontiersinHLTH&FPHY} % for Health, Physics and Mathematics articles
\bibliography{literature}

%%% Make sure to upload the bib file along with the tex file and PDF
%%% Please see the test.bib file for some examples of references

\end{document}